\documentclass[sigconf,nonacm]{acmart}

\usepackage{amsmath,amsthm,bm}
\usepackage{booktabs}
\usepackage{graphicx}
\usepackage{placeins}

\newtheorem{proposition}{Proposition}

\begin{document}
\raggedbottom
\vfuzz=2pt

\title{FunnelCausalNet: Funnel-aware Joint Conversion-Revenue Uplift for Multi-tier Coupon Allocation}

\author{Yu Zhang}
\orcid{0009-0003-5368-1356}
\affiliation{%
  \institution{AMap Alibaba Group}
  \city{Beijing}
  \country{China}}
\email{yuanyu.zy@alibaba-inc.com}

\author{Zhihan Wang}
\orcid{0009-0000-8880-7172}
\affiliation{%
  \institution{AMap Alibaba Group}
  \city{Beijing}
  \country{China}}
\email{yingchen.wzh@alibaba-inc.com}

\author{Guanlin Chen}
\orcid{0009-0008-1374-3926}
\affiliation{%
  \institution{AMap Alibaba Group}
  \city{Beijing}
  \country{China}}
\email{cgl517639@alibaba-inc.com}

\author{Min Jiang}
\orcid{0009-0007-0949-6048}
\affiliation{%
  \institution{AMap Alibaba Group}
  \city{Beijing}
  \country{China}}
\email{jiangmin.jiang@alibaba-inc.com}

\author{Shuai Li}
\affiliation{%
  \institution{AMap Alibaba Group}
  \city{Beijing}
  \country{China}}
\email{lion.lis@alibaba-inc.com}

\renewcommand{\shortauthors}{Zhang et al.}
\hypersetup{%
  pdfauthor={Yu Zhang, Zhihan Wang, Guanlin Chen, Min Jiang, Shuai Li},
  pdfkeywords={uplift modeling, causal inference, heterogeneous treatment effects, coupon allocation, multi-arm randomized experiments, conformal prediction, Lagrangian relaxation, e-commerce}}

\begin{abstract}
Coupon campaigns aim to lift both conversion and revenue, but gross merchandise value (GMV) inherits a deterministic funnel structure from conversion and conditional order value and is typically zero-inflated and heavy-tailed. We propose \emph{FunnelCausalNet}, an uplift estimator that couples a binary conversion head with a nonnegative conditional-value head under the funnel composition $\mu_{\mathrm{gmv}}=\mu_{\mathrm{conv}}\,\mu_{\mathrm{val}}$. Under explicit RCT, support, rate-gap, and cross-head covariance-control assumptions, we derive an idealized leading-order MSE-ratio comparison that identifies a variance regime in which funnel composition can reduce pointwise estimation variance; it is a regime heuristic rather than a guarantee for the shared-representation neural implementation. The estimator is paired with marginal split-conformal summaries on each outcome's CATE (Bonferroni union for two-number joint coverage, treated as audit/monitoring bands) and a Lagrangian budgeted allocator that consumes RCT-anchored estimates for subsidy-aware ROI accounting. On semi-synthetic multi-tier Criteo-MT7 with oracle individualized treatment effects, FunnelCausalNet's mean AUUC\_GMV is within one seed standard deviation of the leading recent feature-interaction baseline among eleven baselines, and a controlled funnel-coupling ablation reduces GMV effect error against direct GMV regression by $18$--$48\%$ across the tested zero-inflation regimes. On de-identified industrial Hotel-Coupon RCT logs with $\approx\!4.9\!\times\!10^{6}$ hold-out exposure records per seed, RCT-consistent expected-outcome (EOM) evaluation sweeps full LP frontiers, and FunnelCausalNet attains the best seed-averaged mean $\Delta\mathrm{ROI}$ at $7/7$ correlated EOM anchors in $10\%$--$60\%$; we treat this as descriptive frontier consistency rather than independent-anchor significance. On sparse binary-spend public benchmarks, revenue-focused rankers can dominate uplift-curve proxies; we foreground this regime boundary explicitly.
\end{abstract}



\maketitle

\section{Introduction}
\label{sec:introduction}

Digital coupon programs aim to lift both the probability that a user converts and the revenue generated conditional on conversion. In practice, marketing teams often estimate conversion uplift and revenue-related quantities through loosely coupled pipelines---separate models for conversion probability and order value---and combine predictions downstream for targeting or budget allocation. This decoupled workflow ignores three structural features that routinely appear in coupon randomized controlled trials (RCTs).

\emph{(i) Funnel identity.} Gross merchandise value (GMV) satisfies $Y^{g}{=}0$ whenever $Y^{c}{=}0$, so GMV decomposes algebraically into conversion mass and conditional spend. Treating GMV as an unconstrained continuous response under extreme zero inflation produces variance-dominated estimates of heterogeneous treatment effects (HTE) on revenue.
\emph{(ii) Ranking divergence.} Ordering users by estimated conversion uplift can disagree substantially with ordering by estimated GMV uplift. Under tight budgets, this inconsistency directly translates into lost incremental GMV relative to revenue-aligned objectives.
\emph{(iii) Multi-tier action space with tier-specific conversion- and revenue-elasticities.} Retailers choose among multiple discount tiers; in our industrial RCT logs different coupon strengths exhibit \emph{quantitatively different} elasticities on conversion probability and on conditional spend---a heavier coupon may convert more users while also reshaping the spend distribution among converters in ways that a single binary lift cannot pin down. Decision-making therefore requires assigning limited subsidy budgets across users \emph{and} arms, and a binary-encouragement formulation cannot answer ``who should receive which tier'' under explicit cost-of-promotion constraints.

These observations motivate \emph{funnel-aware multi-outcome uplift modeling}: jointly estimate heterogeneous causal effects on conversion and GMV while respecting the funnel structure, quantify uncertainty to support conservative deployment, and feed estimates into scalable budgeted multi-tier allocation. We treat the contribution as an end-to-end stack---point estimates, audit-oriented uncertainty, and budgeted assignment with subsidy accounting---rather than a single estimator.

\subsection*{Contributions}
\textbf{(1) Funnel-structured uplift estimation under zero inflation, with a regime-guided variance analysis.} We present \emph{FunnelCausalNet}, an estimator that couples a binary conversion head with a nonnegative spending head consistent with the deterministic zero mass on GMV. Under explicit assumptions on RCT identification, support, convergence rates, and covariance control (Sec.~\ref{sec:method:theory}, Proposition~2), we derive an idealized leading-order MSE-ratio comparison for the high-zero-inflation regime. The comparison provides directional guidance; it does not guarantee dominance for the shared-representation neural implementation or across datasets.
\textbf{(2) Budgeted multi-tier allocation with RCT anchoring.} We combine funnel estimates with Lagrangian relaxation for large-scale multi-tier assignment under subsidy budgets, and absorb additive shifts estimated from RCT arm averages on a held-in slice to mitigate systematic GMV-level bias before forming allocator rewards.
\textbf{(3) Auditable joint uncertainty layer.} We supply marginal split-conformal intervals on each outcome's CATE summary together with a Bonferroni union (a finite-sample valid two-number joint coverage statement) and a Top-$K$ boundary screen flagging users with unstable cross-objective rankings; these are positioned as audit/monitoring bands for compliance review rather than allocator inputs (Sec.~\ref{sec:method:cp}).

\textbf{Empirical scope.} We evaluate against eleven baselines---meta-learners (S-, T-, X-Learner~\cite{kunzel2019metalearners}), causal forests~\cite{wager2018estimation,athey2019generalized}, a dual-head network, CFRNet~\cite{shalit2017estimating}, DragonNet~\cite{shi2019dragonnet}, EFIN~\cite{liu2023efin}, DESCN~\cite{zhong2022descn}, ECUP~\cite{huang2024entire}, and RERUM~\cite{he2024rerum}---on (i) semi-synthetic Criteo-MT7 with oracle individualized treatment effects (ITEs), (ii) the public Hillstrom~\cite{hillstrom2008minethatdata} RCT (3-arm encouragement collapsed to a single send-vs-control contrast as in standard uplift evaluation), (iii) controlled funnel ablations, (iv) joint conformal coverage and computational scaling to $10^{6}$ users, and (v) a large industrial Hotel-Coupon multi-arm RCT with $\approx\!4.9\times10^{6}$ hold-out exposure records per seed under RCT-consistent expected-outcome (EOM) evaluation~\cite{yan2023marketingeom} that sweeps full LP frontiers, with $\Delta\mathrm{GMV\%}$ anchors chosen so realized $\Delta\mathrm{ROI}$ values straddle the break-even band predicted by a commission-rate sensitivity scan (Sec.~\ref{sec:exp:e7}).

\textbf{Honest benchmarking.} Funnel coupling targets multi-tier RCT regimes in which heterogeneous coupon strengths induce \emph{distinct} conversion- and revenue-elasticities. Public benchmarks built around a single send-vs-control encouragement (e.g., Hillstrom) instantiate a \emph{different} decision problem---there is no tier-strength axis along which conversion- and spend-elasticities can differ---so revenue-focused rankers tuned for binary AUUC proxies can score higher. We surface this scope boundary in Section~\ref{sec:discussion} rather than suppressing it. The anchored allocator and the audit conformal layer nonetheless provide a uniform pipeline aligned with subsidy accounting in all regimes.

\section{Related Work}
\label{sec:related}

\paragraph{Uplift modeling and heterogeneous treatment effects.}
Classical uplift estimators include meta-learners (S/T/X) \cite{kunzel2019metalearners}, tree-based CATE estimators such as causal forests \cite{wager2018estimation,athey2019generalized}, R-learner-style residualization \cite{nie2021quasi}, representation-based networks \cite{shalit2017estimating}, and propensity-aware dual-head architectures \cite{shi2019dragonnet}; theoretical generalization guarantees for uplift have also been established under appropriate assumptions \cite{betlei2021uplift}. Surveys synthesize the area \cite{zhang2021unified,devriendt2018literature,gutierrez2017causal}. More recent CIKM work fuses incomplete observational logs with RCT data to identify HTEs when randomized data alone is small \cite{yao2024cio}, complementary to our RCT-anchored funnel composition. Recommender-systems work additionally reformulates top-$N$ recommendation as treatment-effect estimation under exposure-ratio policies \cite{chen2024upliftrec}. Most deployed uplift studies emphasize a single outcome (often conversion) or regress GMV directly without algebraic coupling, which becomes inefficient under dominant zero mass.

\paragraph{Entire-space modeling and deep multi-outcome uplift.}
Post-click conversion-rate estimation widely uses entire-space multi-task objectives that share statistical strength across stages \cite{ma2018esmm,wang2022escm2}. Deep uplift architectures such as DESCN \cite{zhong2022descn} represent treatment heterogeneity over multiple stages, EFIN \cite{liu2023efin} models treatment-aware feature interactions for fine-grained ITE, multi-treatment multi-task uplift addresses tiered responses across arms \cite{wei2024mtmt,zhao2017uplift}, and chain-style models combine awareness--conversion stages with treatment-aware modules \cite{huang2024entire}. Closely related, customer-lifetime-value (CLTV) estimators tackle the heavy-tailed continuous-revenue head with mixture-of-distribution selection \cite{weng2024optdist}, complementary to but distinct from explicit funnel composition. These objectives improve predictive accuracy under multi-task supervision but do not enforce the deterministic funnel link between binary conversion and nonnegative GMV when estimating uplift.

\paragraph{Revenue uplift and budgeted coupon allocation.}
Revenue-focused uplift methods emphasize ranking quality under heavy-tailed continuous outcomes \cite{he2024rerum}. Industrial deployments couple uplift estimates with constrained allocation: real-time coupon allocation cast as a multi-choice knapsack with intent detection \cite{li2020spending}, Lagrangian-style dual updates for real-time coupon allocation \cite{tu2024realtime,kong2026saco}, unified marketing-budget allocation plugging heterogeneous value into constrained assignment \cite{zhao2019unified}, online multi-choice knapsack personalization driven by uplift \cite{albert2022ecommerce}, and end-to-end differentiable allocation for budgeted incentives \cite{sun2024e3ir}. We pair funnel-coupled estimates with RCT anchoring and Lagrangian relaxation, keeping the funnel identity throughout the stack rather than only at the prediction layer.

\paragraph{Uncertainty quantification under treatment effects.}
Conformal prediction yields finite-sample marginal coverage under exchangeability \cite{vovk2005algorithmic}, and conformalized quantile regression tightens intervals for continuous outcomes \cite{roman2019conformalized}. Counterfactual conformal inference extends to ITE-style targets under appropriate sampling designs \cite{lei2021conformal}, and conformal calibration must respect RCT splits to retain causal validity \cite{alaa2019validation}. We compose marginal CQR-style intervals on dual outcomes under a Bonferroni union, producing finite-sample valid two-number coverage statements for joint events rather than a uniformly simultaneous bivariate band.

\paragraph{Hold-out evaluation under RCT logs.}
Yan et al.~\cite{yan2023marketingeom} formalize an expected-outcome metric (EOM) that uses RCT logs to evaluate budgeted policies via H{\'a}jek IPW on policy-matched subsets while sweeping a dual multiplier; we adopt this protocol on the industrial OTA data to obtain a full $(\Delta\mathrm{GMV\%},\Delta\mathrm{ROI})$ frontier rather than a single operating point.

\paragraph{Positioning.}
Deep uplift architectures \cite{zhong2022descn} and revenue-centric estimators \cite{he2024rerum} flexibly model treatment heterogeneity but do not encode the deterministic funnel identity; dual-head networks typically estimate parallel heads without enforcing algebraic consistency between conversion and GMV expectations. These families address different layers of the deployment problem: revenue rankers optimize ordering without enforcing the conversion--GMV support relation, multi-task uplift models do not by themselves provide subsidy-aware allocation, and budgeted incentive methods generally consume effect estimates as plug-ins without RCT arm-level recalibration. Our contribution is the integration of funnel-consistent estimation, audit-oriented dual-outcome summaries, and tier-aware allocation; we do not claim that each constituent mechanism is individually new. Table~\ref{tab:positioning} summarizes the resulting scope differences.

\begin{table}[!htbp]
  \centering\footnotesize
  \caption{Positioning relative to representative method families. ``Funnel'' = enforce $\mu_{\mathrm{gmv}}{=}\mu_{\mathrm{conv}}\mu_{\mathrm{val}}$; ``Joint UQ'' = joint conformal summaries on dual outcomes; ``Multi-tier alloc.'' = native budgeted allocation across $K{>}1$ promotion arms.}
  \label{tab:positioning}
  \resizebox{\columnwidth}{!}{%
  \begin{tabular}{@{}p{2.4cm}cccp{2.6cm}@{}}
    \toprule
    Family & Funnel & Joint UQ & Multi-tier alloc.\ & Coupon-GMV limitation \\
    \midrule
    Meta-learners / forests \cite{kunzel2019metalearners,wager2018estimation} & -- & -- & via post-hoc & Single outcome; ad-hoc composition for GMV. \\
    Entire-space CVR \cite{ma2018esmm,wang2022escm2} & partial & -- & -- & Prediction-targeted, not RCT CATE. \\
    Deep uplift \cite{zhong2022descn,wei2024mtmt,huang2024entire} & -- & -- & native & Parallel heads without algebraic consistency. \\
    Revenue uplift \cite{he2024rerum} & -- & -- & via post-hoc & Ranks GMV; ignores deterministic zeros. \\
    Budgeted incentives \cite{albert2022ecommerce,sun2024e3ir,tu2024realtime,zhao2019unified} & -- & -- & native & Uplift signals plugged in but not coupled. \\
    \textbf{This work} & \textbf{hard} & \textbf{Bonferroni} & \textbf{LP+Lagrange} & Funnel + joint conformal + anchored multi-tier allocation. \\
    \bottomrule
  \end{tabular}}
\end{table}

\section{Problem Formulation}
\label{sec:problem}

\paragraph{Observables and funnel support.}
Consider a coupon RCT with observables $(X,T,Y^{c},Y^{g})$, where $X\in\mathcal{X}$ are user features, $T\in\{0,1,\ldots,K\}$ is a discrete treatment indicator ($T{=}0$ is control), $Y^{c}\in\{0,1\}$ is conversion, and $Y^{g}\in\mathbb{R}_{\ge 0}$ is GMV. Here $T$ indexes the coupon offers randomized in the logs; the method does not interpolate a continuous dose--response curve between those arms. Throughout we impose the \emph{funnel support restriction}
\begin{equation}
\label{eq:funnel-support}
Y^{g}=0 \quad\text{whenever}\quad Y^{c}=0,
\end{equation}
so GMV is undefined as a positive outcome until conversion occurs and conditional spend is only meaningful on the converting subpopulation.

\paragraph{Causal targets.}
Let $Y^{c}(t),Y^{g}(t)$ denote potential outcomes under assignment $t$. For each non-control arm $t\in\{1,\ldots,K\}$, define arm-specific CATEs
\begin{equation}
\label{eq:cate-def}
\tau^{c}_{t}(x)=\mathbb{E}[Y^{c}(t){-}Y^{c}(0)\mid X{=}x],~~
\tau^{g}_{t}(x)=\mathbb{E}[Y^{g}(t){-}Y^{g}(0)\mid X{=}x].
\end{equation}
We identify $\tau^{c}_{t}$ and $\tau^{g}_{t}$ under randomized $T\mid X$ (RCT) as in standard analyses; we do \emph{not} claim identification from purely observational logs.

\paragraph{Tower identity.}
For any fixed arm $t$, the law of iterated expectations gives
\begin{equation}
\label{eq:tower}
\mathbb{E}[Y^{g}(t)]=\mathbb{E}\!\big[\,Y^{c}(t)\cdot\mathbb{E}[Y^{g}(t)\mid Y^{c}(t){=}1,X]\,\big],
\end{equation}
separating \emph{level calibration} of GMV (anchoring metrics in Sec.~\ref{sec:method:alloc}) from \emph{heterogeneous ordering} (PEHE/AUUC in Sec.~\ref{sec:experiments}).

\paragraph{Decision problem: budgeted multi-tier allocation.}
A deterministic policy $\pi:\mathcal{X}\to\{0,\ldots,K\}$ assigns each user to control or one tier. Let $c(x,k)$ denote the predicted incremental subsidy cost of assigning $x$ to tier $k$, derived from the tier-specific coupon terms and the campaign-specific accounting base. Feasible policies satisfy a total budget $B{>}0$:
\begin{equation}
\label{eq:budget}
\textstyle\sum_{i} c(x_{i},\pi(x_{i}))\le B,~~ \pi(x_{i})\in\{0,\ldots,K\}.
\end{equation}
Objectives include maximizing incremental GMV
$\textstyle\sum_{i}\mathbb{E}[\tau^{g}_{\pi(x_{i})}(x_{i})]$
or its ROI-style surrogate
$\Delta\mathrm{ROI}{:=}\sum_{i}\hat\tau^{g}_{\pi(x_{i})}(x_{i}) / \sum_{i}\hat c_{i,\pi(x_{i})}$.

\paragraph{Dual-objective tension.}
When rankings induced by $\tau^{c}$ and $\tau^{g}$ disagree, no single scalar objective is universally aligned with business preferences. The conflict diagnostic in Sec.~\ref{sec:method:cp} does \emph{not} solve a general multi-objective program; it flags individuals whose objective-wise rankings and intervals jointly indicate instability near budgeted Top-$K$ cuts.

\section{Methodology}
\label{sec:method}

\subsection{Funnel-structured uplift estimation}
\label{sec:method:funnel}

We estimate multi-arm conversion probabilities $\mu_{\mathrm{conv}}^{(t)}(x)$ and nonnegative conditional order-value expectations $\mu_{\mathrm{val}}^{(t)}(x)$ with shared representations. GMV expectations obey the \emph{funnel composition}
\begin{equation}
\label{eq:funnel-composition}
\mu_{\mathrm{gmv}}^{(t)}(x)=\mu_{\mathrm{conv}}^{(t)}(x)\,\mu_{\mathrm{val}}^{(t)}(x)
\end{equation}
after numerical stabilization (clipping, nonnegative projections). Training combines Bernoulli conversion losses with squared error on $\log(1+\mathrm{GMV})$ among converters; inference maps normalized logits back to currency units using a LogNormal-style mean correction. The total objective allows optional consistency and monotonicity terms:
\begin{equation}
\label{eq:loss-sketch}
\mathcal{L}_{\mathrm{total}}=\mathcal{L}_{\mathrm{conv}}+\alpha\,\mathcal{L}_{\mathrm{val}}+\beta\,\mathcal{L}_{\mathrm{consist}}+\gamma\,\mathcal{L}_{\mathrm{mono}}.
\end{equation}
``Soft funnel'' variants replace the hard product~\eqref{eq:funnel-composition} with large penalties. They can be preferable when stage labels are asynchronously logged, missing, or otherwise make the support relation approximate. In our RCT logs the support identity is verified by construction, and Sec.~\ref{sec:exp:e2} shows that retaining violations through a soft penalty does not match hard composition under extreme zero inflation.

\subsection{Variance decomposition and leading-order MSE ratio}
\label{sec:method:theory}

Fix $(X,T){=}(x,t)$. For the following propositions, all moments are conditional on this event; let $p{:=}\mathbb{E}[Y^{c}]$, $\mu_v{:=}\mathbb{E}[Y^{g}\mid Y^{c}{=}1]$, and $\sigma_v^{2}{:=}\mathrm{Var}(Y^{g}\mid Y^{c}{=}1)$.

\begin{proposition}[Variance decomposition]\label{prop:var-decomp}
\begin{equation}
\label{eq:var-decomp}
\mathrm{Var}(Y^{g}\mid X{=}x,T{=}t)=p\,\sigma_v^{2}+p(1-p)\,\mu_v^{2}.
\end{equation}
The first term is the within-converter variance; the second is the Bernoulli switching variance contributed by the zero mass.
\end{proposition}

\begin{proposition}[Idealized leading-order MSE ratio under a rate gap]\label{prop:mse-ratio}
Let $\hat{\mu}_g^{\mathrm{direct}}$ be a direct nonparametric squared-error estimator of $\mathbb{E}[Y^{g}\mid X{=}x,T{=}t]$ and $\hat{\mu}_g^{\mathrm{funnel}}{:=}\hat{\mu}_{\mathrm{conv}}\hat{\mu}_{\mathrm{val}}$ the funnel composition estimator. Assume:
\begin{itemize}\itemsep 0pt
\item[(A1)] (\emph{RCT identification.}) $T\!\perp\!(Y^{c}(\cdot),Y^{g}(\cdot))\mid X$ and the propensity $\Pr(T{=}t\mid X)$ is bounded away from $0$ at $x$.
\item[(A2)] (\emph{Funnel support.}) $Y^{g}{=}0$ whenever $Y^{c}{=}0$, with $\sigma_v^{2}\!\in\!(0,\infty)$ and $\mu_v\!\in\!(0,\infty)$.
\item[(A3)] (\emph{Conv-head parametric rate.}) $\hat{\mu}_{\mathrm{conv}}$ is fit by a (correctly-specified) parametric Bernoulli model on the full $n$-sample, so $\hat{p}{-}p{=}O_{p}(n^{-1/2})$ at $(x,t)$.
\item[(A4)] (\emph{Value-head and direct nonparametric variance.}) $\hat{\mu}_{\mathrm{val}}$ is fit nonparametrically on the converter subsample and $\hat{\mu}_g^{\mathrm{direct}}$ is fit nonparametrically on the full sample, both with negligible bias under standard undersmoothing and asymptotic pointwise variances scaling as $1/r_n$ for an effective-sample-size sequence $r_n\!\to\!\infty$ with $r_n=o(n)$: $\mathrm{Var}(\hat{\mu}_{\mathrm{val}})\!\sim\!\sigma_v^{2}/(p\,r_n)$ and $\mathrm{Var}(\hat{\mu}_g^{\mathrm{direct}})\!\sim\!\mathrm{Var}(Y^{g}\mid X{=}x,T{=}t)/r_n$.
\item[(A5)] (\emph{Cross-head covariance control.}) Either independent sample splitting makes $\mathrm{Cov}(\hat{p},\hat{\mu}_v){=}0$ in the idealized analysis~\cite{chernozhukov2018double}, or the covariance is $o(r_n^{-1})$. This condition is not guaranteed by a shared-representation neural implementation.
\end{itemize}
Then, applying the delta method to $(p,\mu_v)\!\mapsto\!p\mu_v$, the leading-order pointwise MSEs satisfy
\begin{equation}
\label{eq:mse-ratio}
\lim_{n\to\infty}\frac{\mathrm{MSE}(\hat{\mu}_g^{\mathrm{funnel}})}{\mathrm{MSE}(\hat{\mu}_g^{\mathrm{direct}})}
=\frac{p\,\sigma_v^{2}}{p\,\sigma_v^{2}+p(1-p)\,\mu_v^{2}}
=\frac{1}{1+(1-p)\,\mu_v^{2}/\sigma_v^{2}}.
\end{equation}
A proof sketch is given in Appendix~\ref{app:proof}.
\end{proposition}

\noindent\textit{Sufficient-regime reading.}
Eq.~\eqref{eq:mse-ratio} is an idealized pointwise variance comparison, not a universal optimality theorem or a guarantee for CATE ranking. It relies on the parametric rate gap (A3) and covariance control (A5), which make the Bernoulli switching and cross-head terms vanish faster than the within-converter contribution $p\,\sigma_v^{2}/r_{n}$. If both heads are estimated nonparametrically at the same rate, the ratio collapses to one. Shared neural representations can also induce correlated finite-sample errors, and systematic biases in the two heads can be multiplied by the product composition. We therefore use~\eqref{eq:mse-ratio} only as a regime indicator; the controlled E2 stress test (Sec.~\ref{sec:exp:e2}) probes whether its predicted direction appears empirically across $\hat{p}\!\in\![5,45]\%$ without validating the asymptotic assumptions.

\noindent\textit{Operational regime.}
Within these idealized assumptions, the ratio in~\eqref{eq:mse-ratio} is below one whenever $(1{-}p)\mu_v^{2}/\sigma_v^{2}{>}0$, and shrinks as the zero mass $(1{-}p)$ grows or $\mu_v$ dominates $\sigma_v$. This is the same hurdle/two-part structural intuition long studied in econometrics~\cite{cragg1971some,mullahy1986specification,lambert1992zero}; our contribution is to connect the leading-order ratio under the rate-gap regime to the coupon-uplift setting, not to claim a new funnel identity or universal dominance.

\smallskip
\noindent\textbf{Remark (Bernoulli--LogNormal likelihood alignment).}
When focal weights are inactive, the LogNormal dispersion in the converting subsample is fixed during value-head fitting, and a log-domain Gaussian surrogate matches the implementation's $\log(1{+}\mathrm{GMV})$ regression among converters, the loss $\mathcal{L}_{\mathrm{conv}}{+}\alpha\,\mathcal{L}_{\mathrm{val}}$ coincides with the negative log-likelihood of a hierarchical Bernoulli--LogNormal model for $(Y^{c},Y^{g})$ up to additive constants depending only on hyperparameters. The default hard-mode training path (BCE-with-logits $+$ converter MSE on normalized $\log(1{+}\mathrm{GMV})$) instantiates this idealized limit; the \texttt{ziln} variant swaps in focal-BCE and explicit LogNormal NLL with typically frozen dispersion (Sec.~\ref{sec:exp:e2}). Auxiliary monotonicity and consistency losses are outside this alignment.

\subsection{Joint conformal intervals and conflict screening}
\label{sec:method:cp}

\textbf{Scope.} We compose \emph{marginal} split-conformal intervals on each outcome's CATE summary and apply a Bonferroni union across the two margins. Under standard split-conformal assumptions on a disjoint calibration fold, both intervals jointly cover their respective targets with probability at least $1{-}\alpha$ at nominal level $\alpha/2$ per margin. This is a finite-sample valid \emph{two-number coverage} statement, not a simultaneous band over the bivariate CATE surface.

\textbf{Deployment stance.} In practice, Bonferroni splits, finite-sample CQR offsets, and heavy-tailed GMV residuals drive empirical joint coverage \emph{above} the nominal $1{-}\alpha$ (Sec.~\ref{sec:exp:e5}). We therefore treat intervals primarily as \textbf{auditable monitoring bands} for compliance and risk review, recommend wider nominal $\alpha\in[0.10,0.20]$ when widths must remain actionable, and pair intervals with anchored point estimates when feeding optimizers, because marginally valid lower-conformal bounds for $\tau^{g}$ at narrow $\alpha$ can be so pessimistic under zero inflation that budgeted LCB policies collapse to all-control assignments (Sec.~\ref{sec:exp:e4}).

\textbf{Conflict diagnostic.} A Top-$K$ boundary screen flags users with (i) disagreement between $\tau^{c}$ and $\tau^{g}$ rankings, (ii) wide dual intervals or predictions near decision cutoffs, and (iii) instability near budgeted thresholds. The rule is a risk-disclosure layer for manual review or conservative assignment, \emph{not} a precision-calibrated detector of latent business conflicts.

\subsection{Budgeted multi-tier allocation with anchoring}
\label{sec:method:alloc}

Given predicted incremental rewards $\hat{\tau}^{g}_{t}(x)$ and costs $c(x,t)$, we maximize the budgeted assignment using \textbf{Lagrangian relaxation}: dual updates over a scalar multiplier $\lambda$ approximately enforce~\eqref{eq:budget} while inner problems decouple across users for scalability. Moderate-scale LP relaxations serve as references where memory permits and provide the headline industrial allocator in Sec.~\ref{sec:exp:e7}.

\textbf{Anchoring.} Predicted GMV levels can exhibit systematic bias under extreme zero inflation (for example, underestimating control-arm GMV mass). We apply additive shifts estimated from RCT arm-wise averages on a held-in slice before forming rewards fed to the allocator, improving $\Delta\mathrm{ROI}$-style objectives without retraining. End-to-end anchor losses are left for future work.

\textbf{Train--calibrate--test.} We use disjoint splits: training folds for model fitting, a separate calibration fold for conformal quantiles, and held-out test folds for uplift metrics and allocation summaries. User-level clustering is recommended when the same customer could otherwise leak across folds.

\section{Experiments}
\label{sec:experiments}

\subsection{Datasets and protocol}
\label{sec:exp:protocol}

We compare twelve methods: FunnelCausalNet plus eleven baselines spanning meta-learners (S/T/X)~\cite{kunzel2019metalearners}, causal forests~\cite{wager2018estimation,athey2019generalized}, a dual-head network, CFRNet representation balancing~\cite{shalit2017estimating}, DragonNet propensity-aware twin-head~\cite{shi2019dragonnet}, EFIN explicit feature--treatment interaction~\cite{liu2023efin}, DESCN-style~\cite{zhong2022descn} and ECUP-style~\cite{huang2024entire} deep uplift, and RERUM-style~\cite{he2024rerum} revenue ranking uplift. Multi-tier extensions of binary-treatment originals (CFRNet, DragonNet) follow common practice~\cite{zhao2017uplift,wei2024mtmt}: per-arm outcome heads with the binary balancing/propensity penalty replaced by a multi-arm aggregation (mean pairwise linear-MMD against control for CFRNet; multi-class softmax cross-entropy for DragonNet); EFIN keeps its intent-attention block plus per-arm explicit feature-treatment cross-interaction. All deep baselines are reimplemented under a unified PyTorch pipeline so that training schedules, hyperparameters, and evaluation interfaces are identical across methods.\footnote{A unified harness keeps AUUC, PEHE, and EOM evaluation interfaces comparable across methods; we also perform head-to-head sanity checks against published Hillstrom and Criteo-Uplift results to bound implementation gap.} Models train for $25$ epochs with Adam; uplift metrics aggregate three seeds. Fixed experiment configurations and seeds are used throughout; internal reruns reproduce the reported aggregates up to floating-point nondeterminism.

\begin{table}[!htbp]
  \centering\footnotesize
  \caption{Datasets used in the main matrix.}
  \label{tab:datasets}
  \resizebox{\columnwidth}{!}{%
  \begin{tabular}{@{}llll@{}}
    \toprule
    Dataset & $N$ (default) & Arms & Evaluation \\
    \midrule
    Criteo-MT7 (semi-synth.) & $10\mathrm{K}$ tr./eval & $8{+}1$ & Oracle ITE; PEHE, AUUC \\
    Hillstrom~\cite{hillstrom2008minethatdata}   & $\sim 64\mathrm{K}$  & Binary & AUUC proxies; no oracle PEHE \\
    OTA Hotel-Coupon (de-id.) & $5\mathrm{M}$ records; $50\mathrm{K}/4.9\mathrm{M}$ tr./eval & Multi-tier & RCT EOM \\
    \bottomrule
  \end{tabular}}
\end{table}

\paragraph{Semi-synthetic calibration disclosure.}
Criteo-MT7's generator parameters (baseline conversion ${\approx}8\%$, eight tiers $0\%$--$14\%$) fall inside operationally common e-commerce coupon ranges and are \emph{not} tuned to match any specific industrial dataset. To rule out calibration that selectively favors funnel composition, the E2b stress test (Sec.~\ref{sec:exp:e2}) sweeps the conversion baseline across $\hat{p}\in[4.6\%,45.4\%]$; the funnel benefit is monotone throughout this range.

\paragraph{Public benchmark coverage.}
We surveyed the public corpora cited across the closest baselines (Sec.~\ref{sec:exp:protocol}) and across multi-treatment ITE methodology papers (drawing on the e-commerce uplift surveys of~\cite{devriendt2018literature,gutierrez2017causal}), including MEMENTO~\cite{mondal2022memento}, whose own experiments rely on Amazon-private and fully synthetic data because no public multi-treatment RCT was available to its authors. HTE classics from medical or educational RCTs (IHDP~\cite{hill2011bayesian}, ACIC, Mindsets, TWINS; surveyed in~\cite{devriendt2018literature}) pair binary treatment with continuous outcomes; multi-arm semi-synthetic surfaces (News, TCGA) pair tiered treatments with simulated outcomes over text or genomic covariates rather than coupon RCTs; non-commercial multi-arm RCTs from political (Gerber et al.'s GOTV, 5 arms) or clinical (Colon, 3 arms) trials are likewise scope-mismatched with the coupon-strength setting. E-commerce uplift releases---Hillstrom~\cite{hillstrom2008minethatdata}, Lenta~\cite{lentauplift2020}, MegaFon~\cite{megafonuplift2021}, Criteo Uplift v2.x~\cite{diemert2018large}, and the DESCN-companion Lazada release~\cite{zhong2022descn}---either implement single-encouragement send-vs-control RCT designs or, in Hillstrom's three-arm form, contrast different message \emph{types} (men vs.~women catalog) rather than coupon-strength tiers, so the tier-specific conversion- and revenue-elasticities motivation~(iii) of Sec.~\ref{sec:introduction} is not realized. Multi-arm public collections such as Tianchi-O2O~\cite{tianchio2o2018} provide tiered discount rates but only coupon-redemption labels with no continuous GMV supervision and are observational rather than randomized; the Open Bandit Dataset~\cite{saito2021open} is a recommendation-policy log rather than a coupon-strength RCT. The closest publicly discussed multi-tier coupon RCT is the MT-LIFT release shipped with ECUP~\cite{huang2024entire} (5-arm Meituan food-delivery, ${\approx}5.5\mathrm{M}$ samples), but it provides only binary chain labels (click and conversion) and no continuous-spend / GMV outcome, so the within-converter funnel-value head this paper targets cannot be supervised on it. To our knowledge, no public benchmark simultaneously realizes multi-tier coupon assignment, continuous GMV ground truth, and strict RCT randomization. We therefore evaluate on (a) a semi-synthetic multi-tier surface (Criteo-MT7, oracle ITE), (b) one widely cited public RCT (Hillstrom, included as a scope-boundary disclosure), and (c) a large industrial multi-tier RCT (OTA Hotel-Coupon) that instantiates the target regime at production scale. The implementation uses a modular ingestion interface so that additional multi-tier corpora can be evaluated without changing the model or evaluation logic.

\noindent\emph{Metrics.} \textbf{AUUC\_GMV} and \textbf{AUUC\_CVR} summarize uplift-curve area (higher is better). \textbf{PEHE\_GMV} and \textbf{PEHE\_CVR} are defined where identifiable ground truth is available (Criteo-MT7). \textbf{ATE error} measures GMV average-treatment-effect error when defined. Hillstrom and OTA do not admit the same PEHE as MT7; we report ranking and calibration metrics appropriate to each source. The industrial protocol additionally reports the \textbf{expected-outcome metric (EOM)} of~\cite{yan2023marketingeom} via H{\'a}jek IPW on policy-matched RCT subsets while sweeping a dual multiplier $\alpha$ to trace the $(\Delta\mathrm{GMV\%},\Delta\mathrm{ROI})$ frontier.

\subsection{Uplift estimation quality (E1)}
\label{sec:exp:e1}

Table~\ref{tab:e1-mt7} reports three-seed means on Criteo-MT7 at $N{=}10\mathrm{K}$. EFIN attains the highest AUUC\_GMV ($0.615$); FunnelCausalNet ranks second ($0.613$, within one seed standard deviation), indicating that EFIN's explicit feature-treatment cross-interaction aligns particularly well with this synthetic generator's tier-aware nonlinearity. Crucially, the semi-synthetic advantage does \emph{not} carry over to the production OTA RCT (Sec.~\ref{sec:exp:e7}), where FunnelCausalNet leads at every $\Delta\mathrm{GMV\%}$ anchor. \textbf{PEHE\_CVR} is led by DualHeadNet ($0.048$); FunnelCausalNet ($0.058$) remains competitive, confirming that funnel coupling does not destroy conversion-head identifiability. \textbf{ATE\_GMV\_err} favors shallower models (Causal Forest, S-Learner) that compress the conditional-mean range---level calibration and heterogeneous ranking are distinct objectives.

\begin{table}[!htbp]
  \centering\footnotesize
  \caption{Criteo-MT7 estimation quality (E1; $N{=}10\mathrm{K}$, three-seed means). Arrows indicate desired direction; boldface marks the best mean per column.}
  \label{tab:e1-mt7}
  \resizebox{\columnwidth}{!}{%
  \begin{tabular}{@{}lrrrr@{}}
    \toprule
    Method & AUUC\_GMV $\uparrow$ & PEHE\_GMV $\downarrow$ & PEHE\_CVR $\downarrow$ & ATE\_err $\downarrow$ \\
    \midrule
    EFIN                     & \textbf{0.615} & \textbf{26.70} & 0.054 & 12.41 \\
    FunnelCausalNet          & 0.613 & 31.18 & 0.058 & 21.67 \\
    DESCN-style              & 0.605 & 30.91 & 0.054 & 20.88 \\
    DragonNet                & 0.593 & 27.96 & 0.056 & 14.47 \\
    CFRNet                   & 0.568 & 29.29 & 0.064 & 14.06 \\
    ECUP                     & 0.567 & 32.39 & 0.055 & 22.51 \\
    RERUM                    & 0.562 & 39.17 & 0.072 & 28.75 \\
    DualHeadNet              & 0.514 & 32.44 & \textbf{0.048} & 7.43 \\
    S-Learner                & 0.508 & 43.71 & 0.067 & 5.98 \\
    X-Learner                & 0.508 & 148.7 & 0.136 & 8.14 \\
    T-Learner                & 0.507 & 197.5 & 0.182 & 6.02 \\
    Causal Forest            & 0.489 & 56.73 & 0.080 & \textbf{4.62} \\
    \bottomrule
  \end{tabular}}
\end{table}

\noindent\emph{Public-RCT scope boundary.} Hillstrom is a single-encouragement RCT in which two message-type arms are pooled against the no-send control; positives are sparse and there is no coupon-strength axis along which conversion- and spend-elasticities can differ. The multi-tier decision problem this paper targets (Sec.~\ref{sec:introduction}, contribution~(iii)) is therefore not realized, and the funnel composition reduces to estimating a near-degenerate spending head downstream of a single conversion lift. With extremely sparse converters, the value head has too few effective samples to outperform a direct rank-style estimator on revenue. Empirically, revenue-focused rankers RERUM ($0.747$) and DualHeadNet ($0.739$) lead AUUC\_GMV on Hillstrom, while \emph{all} multi-tier funnel-aware deep models (DESCN, ECUP, FunnelCausalNet) underperform. This result may reflect both scope mismatch and a finite-sample converter bottleneck, and it is direct evidence that FunnelCausalNet is not broadly dominant on binary public benchmarks. The industrial OTA multi-arm RCT in Sec.~\ref{sec:exp:e7} is the target regime rather than proof of transfer beyond it.

\subsection{Funnel ablation (E2)}
\label{sec:exp:e2}

We compare four modes on Criteo-MT7: direct GMV regression (A), soft funnel penalties (B), hard funnel composition (C), and a ZILN-style likelihood path (D), using five seeds for each sample size and mode. Hard coupling achieves the lowest PEHE\_GMV at $10\mathrm{K}$, $20\mathrm{K}$, and $100\mathrm{K}$ samples, while the funnel-violation rate of A remains at $\gtrsim 60\%$ versus $0\%$ for C. At $N{=}100\mathrm{K}$, D approaches C ($17.7$ vs.\ $16.0$; relative excess $\approx 11\%$), consistent with the Bernoulli$\times$LogNormal narrative; at $N{=}10\mathrm{K}$, D underperforms hard composition because of limited converter sample for fitting the alternate likelihood. This ablation isolates the funnel formulation while holding the training harness fixed; E4 separately compares allocation variants. We do not claim a full factorial decomposition of estimator architecture, anchoring, conformal diagnostics, and allocation.

\begin{table}[!htbp]
  \centering\footnotesize
  \caption{E2 Criteo-MT7 funnel ablation: mean PEHE\_GMV (lower better) and funnel-violation rate (\%, 5 seeds). Modes: A=direct GMV regression, B=soft penalty, C=hard funnel, D=ZILN-style.}
  \label{tab:e2}
  \resizebox{\columnwidth}{!}{%
  \begin{tabular}{@{}rcccc@{\hskip 6pt}cccc@{}}
    \toprule
    & \multicolumn{4}{c}{PEHE\_GMV $\downarrow$} & \multicolumn{4}{c}{Violation (\%) $\downarrow$} \\
    \cmidrule(lr){2-5}\cmidrule(l){6-9}
    $N$ & A & B & C & D & A & B & C & D \\
    \midrule
    $10\mathrm{K}$  & 25.94 & 26.12 & \textbf{20.26} & 38.02 & 62.4 & 1.21 & \textbf{0.00} & 0.00 \\
    $20\mathrm{K}$  & 25.48 & 25.91 & \textbf{15.29} & 23.23 & 64.0 & 1.34 & \textbf{0.00} & 0.00 \\
    $100\mathrm{K}$ & 25.80 & 25.43 & \textbf{15.97} & 17.67 & 71.2 & 1.25 & \textbf{0.00} & 0.00 \\
    \bottomrule
  \end{tabular}}
\end{table}

\noindent\emph{Prop.~2 zero-inflation stress test.}
We probe the variance regime suggested by~\eqref{eq:mse-ratio} on a controlled semi-synthetic surface by sweeping the conversion baseline logit $\mu_{p}$ of the Criteo-MT7 generator at $N{=}20\mathrm{K}$, $5$ seeds each, contrasting hard funnel composition (C\_hard) against direct GMV regression (A\_direct). Table~\ref{tab:e2b-prop2} reports observed conversion rate $\hat{p}$ and PEHE\_GMV ratio across four levels. Funnel composition reduces PEHE\_GMV by $18$--$48\%$ across the tested $\hat{p}\in[4.6\%,45.4\%]$ range, with peak benefit at moderate-high zero inflation; this direction is consistent with Eq.~\eqref{eq:mse-ratio}, but the ablation does not verify its asymptotic assumptions or establish general dominance. The finite-sample dip at the most extreme $\hat{p}\!=\!4.6\%$ end is consistent with sparse-converter noise on the value head.

\begin{table}[!htbp]
  \centering\scriptsize
  \caption{E2 extension: Prop.~2 zero-inflation stress test on Criteo-MT7 ($N{=}20\mathrm{K}$, $5$ seeds). PEHE\_GMV mean$\pm$std for direct GMV regression (A) versus hard funnel composition (C); benefit$=1{-}\mathrm{PEHE}_{C}/\mathrm{PEHE}_{A}$. All four tested levels favor C, with peak benefit at moderate-high zero inflation, consistent with the direction of~\eqref{eq:mse-ratio}.}
  \label{tab:e2b-prop2}
  \begin{tabular}{@{}lcccc@{}}
    \toprule
    Level $\mu_{p}$ & $\hat{p}$ & PEHE\_GMV (A) & PEHE\_GMV (C) & Benefit \\
    \midrule
    $-3.5$ & $4.6\%$  & $12.22{\pm}3.35$ & $\mathbf{10.01}{\pm}3.57$ & $+18.1\%$ \\
    $-2.4$ & $11.9\%$ & $25.43{\pm}6.25$ & $\mathbf{15.77}{\pm}5.79$ & $+38.0\%$ \\
    $-1.5$ & $24.3\%$ & $39.22{\pm}8.96$ & $\mathbf{20.29}{\pm}5.70$ & $\mathbf{+48.3\%}$ \\
    $-0.5$ & $45.4\%$ & $55.11{\pm}15.91$ & $\mathbf{30.54}{\pm}5.65$ & $+44.6\%$ \\
    \bottomrule
  \end{tabular}
\end{table}

\subsection{Conflict diagnostic as audit layer (E3)}
\label{sec:exp:e3}

This subsection sanity-checks the Top-$K$ conflict screen of Sec.~\ref{sec:method:cp} as an \emph{audit signal}, not as a production classifier; it is not part of the funnel-estimation or budgeted-allocation pipelines. We use synthetic stress tests with controllable injection (no latent conflict labels exist on real coupon logs), varying an injection correlation $\rho_{\mathrm{conf}}$ between conversion and GMV uplift signals. Table~\ref{tab:e3-conflict} reports three-seed means of precision, recall, and F$_1$. Peak F$_1$ reaches $\approx 0.25$ at $\rho_{\mathrm{conf}}{=}0.6$; the rule's value is to surface users near budget cuts whose objective-wise orderings disagree, not to act as a calibrated detector.

\begin{table}[!htbp]
  \centering\footnotesize
  \caption{E3 semi-synthetic conflict detection: precision / recall / F$_1$ versus injected correlation $\rho_{\mathrm{conf}}$ (three-seed means).}
  \label{tab:e3-conflict}
  \begin{tabular}{@{}rccc@{}}
    \toprule
    $\rho_{\mathrm{conf}}$ & Precision & Recall & F$_1$ \\
    \midrule
    $0.0$ & 0.146 & 0.116 & 0.120 \\
    $0.3$ & 0.201 & 0.175 & 0.179 \\
    $0.6$ & 0.286 & 0.218 & \textbf{0.246} \\
    $0.9$ & 0.203 & 0.116 & 0.146 \\
    \bottomrule
  \end{tabular}
\end{table}

\subsection{Budgeted allocation on MT7 (E4)}
\label{sec:exp:e4}

On semi-synthetic Criteo-MT7 at $N{=}20\mathrm{K}$ with eight tiers and realistic cost presets, we pipeline FunnelCausalNet predictions through joint conformal summaries (where applicable) and budgeted allocation. With tier discount rates $d_k$, the solver uses costs $d_k\hat\mu_g^{(k)}(x)$, while evaluation applies the same rule to the oracle GMV surface. Table~\ref{tab:e4-mt7-roi} reports the mean oracle incremental-GMV surrogate $\tau_g$, realized subsidy cost, and their ratio $\Delta\mathrm{ROI}$ across three seeds.

The \textbf{anchored-Lagrangian pipeline} attains higher $\Delta\mathrm{ROI}$ than random allocation under tight budgets---for example, $3.92$ versus $3.07$ at $B/B_{\mathrm{free}}{=}0.05$---with lower realized cost and competitive incremental GMV. This comparison changes anchoring and allocation jointly and therefore does not isolate the anchoring contribution. \textbf{LP} relaxation often achieves higher raw incremental GMV but spends more budget; at $B/B_{\mathrm{free}}{=}0.50$ the anchored-Lagrangian pipeline tracks alternatives on $\Delta\mathrm{ROI}$ while trading off peak GMV. \textbf{LCB}-driven assignments (\texttt{funnel\_ip\_lcb}) degenerate to all-control allocations in these logs (zero realized lift and cost), consistent with pessimistic lower-conformal surfaces under heavy zero inflation; they are omitted from Table~\ref{tab:e4-mt7-roi}, motivating the deployment stance in Sec.~\ref{sec:method:cp} (wider $\alpha$ or anchored point estimates rather than narrow-$\alpha$ LCB).

\begin{table}[!htbp]
  \centering\footnotesize
  \caption{E4 semi-synthetic MT7 budgeted allocation versus baselines (three-seed means). $\Delta\mathrm{ROI}=$(oracle incremental GMV)/(realized subsidy cost).}
  \label{tab:e4-mt7-roi}
  \resizebox{\columnwidth}{!}{%
  \begin{tabular}{@{}rlrrr@{}}
    \toprule
    $B/B_{\mathrm{free}}$ & Strategy & $\tau_g$ surr.\ & Cost & $\Delta\mathrm{ROI}$ \\
    \midrule
    $0.05$ & \texttt{baseline\_random}       & 8\,783 & 2\,864 & 3.07 \\
    $0.05$ & \texttt{baseline\_topk}         & 8\,813 & 2\,409 & 3.66 \\
    $0.05$ & \texttt{funnel\_ip\_lp}         & 9\,986 & 2\,910 & 3.43 \\
    $0.05$ & \texttt{funnel\_ip\_anchored}   & 9\,072 & 2\,315 & \textbf{3.92} \\
    \midrule
    $0.10$ & \texttt{baseline\_random}       & 16\,460 & 5\,580 & 2.95 \\
    $0.10$ & \texttt{funnel\_ip\_anchored}   & 16\,379 & 4\,413 & \textbf{3.71} \\
    \midrule
    $0.50$ & \texttt{baseline\_topk}         & 71\,906 & 23\,593 & 3.05 \\
    $0.50$ & \texttt{funnel\_ip\_lp}         & 72\,870 & 24\,069 & 3.03 \\
    $0.50$ & \texttt{funnel\_ip\_anchored}   & 63\,766 & 21\,254 & 3.00 \\
    \bottomrule
  \end{tabular}}
\end{table}

\subsection{Joint conformal coverage (E5)}
\label{sec:exp:e5}

We run split conformal with Bonferroni separation across conversion and GMV outcomes using three seeds on MT7 ($N{=}20\mathrm{K}$) and OTA ($N{=}50\mathrm{K}$). Table~\ref{tab:e5-y-layer} (left block) reports marginal and joint outcome-layer coverage; the right block of the same table quantifies systematic over-coverage of the joint event relative to nominal $1{-}\alpha$. Table~\ref{tab:e5-tau-mt7} lists oracle-$\tau$ coverage on MT7 (identifiable) together with mean $\tau_g$ interval width in currency units.

\begin{table}[!htbp]
  \centering\scriptsize
  \caption{E5 outcome-layer empirical coverage (three-seed means; rounded).}
  \label{tab:e5-y-layer}
  \begin{tabular}{@{}llccc@{}}
    \toprule
    Dataset & $\alpha$ & cov\_conv & cov\_val($c{=}1$) & cov\_joint \\
    \midrule
    MT7 & 0.05 & 0.989 & 0.986 & 0.987 \\
    MT7 & 0.10 & 0.975 & 0.986 & 0.974 \\
    MT7 & 0.20 & 0.947 & 0.949 & 0.942 \\
    OTA & 0.05 & 0.989 & 0.986 & 0.988 \\
    OTA & 0.10 & 0.975 & 0.980 & 0.974 \\
    OTA & 0.20 & 0.950 & 0.959 & 0.948 \\
    \bottomrule
  \end{tabular}\hfill
  \begin{tabular}{@{}rccc@{}}
    \toprule
    $\alpha$ & nominal & MT7 $\Delta$ & OTA $\Delta$ \\
    \midrule
    $0.05$ & 0.95 & $+0.037$ & $+0.038$ \\
    $0.10$ & 0.90 & $+0.074$ & $+0.074$ \\
    $0.20$ & 0.80 & $+0.142$ & $+0.148$ \\
    \multicolumn{4}{l}{\scriptsize $\Delta = $ cov\_joint $-(1{-}\alpha)$ in pp.} \\
    \bottomrule
  \end{tabular}
\end{table}

\begin{table}[!htbp]
  \centering\footnotesize
  \caption{E5 MT7 oracle-$\tau$ coverage and mean $\tau_g$ interval width (currency units, three-seed means). OTA $\tau$-oracle summaries omitted.}
  \label{tab:e5-tau-mt7}
  \begin{tabular}{@{}rcccc@{}}
    \toprule
    $\alpha$ & cov $\tau_c$ & cov $\tau_g$ & Mean $w_{\tau_g}$ & Mean $w_{\tau_c}$ \\
    \midrule
    $0.05$ & 1.000 & 1.000 & 11\,741 & 2.000 \\
    $0.10$ & 1.000 & 1.000 & 11\,637 & 2.000 \\
    $0.20$ & 1.000 & 1.000 & 8\,748  & 2.000 \\
    \bottomrule
  \end{tabular}
\end{table}

\begin{figure}[!htbp]
  \centering
  \includegraphics[width=\linewidth]{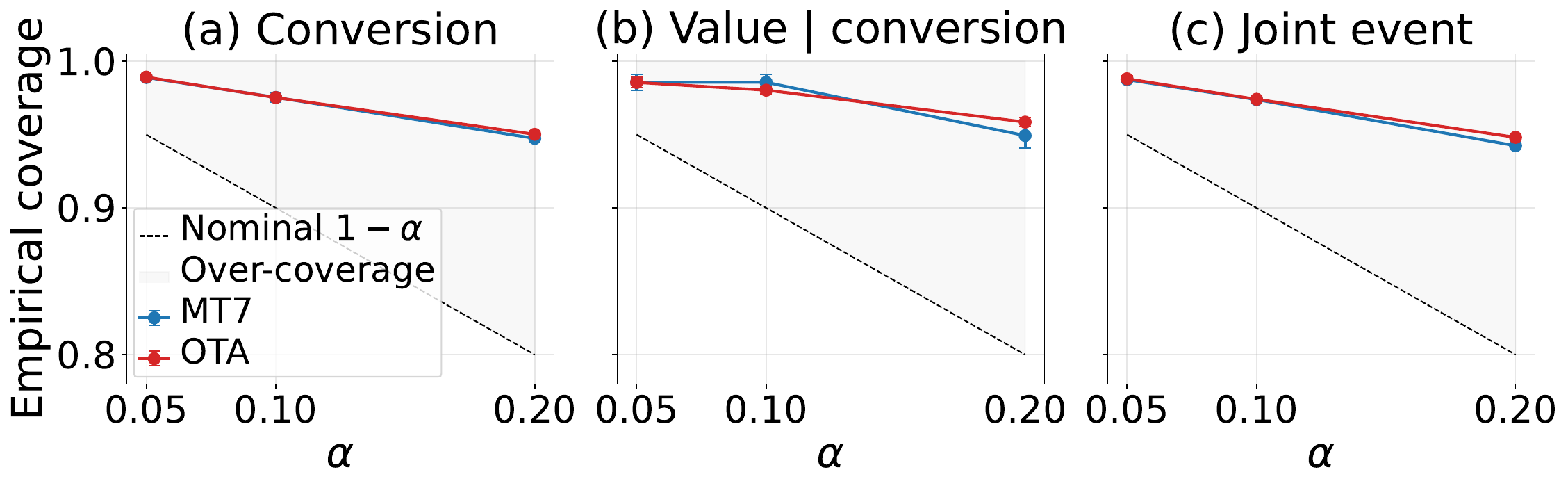}
  \Description{Bar chart of empirical outcome-layer conformal coverage versus nominal alpha for MT7 and OTA datasets. For each dataset and alpha in 0.05/0.10/0.20, three bars show marginal conversion coverage, conditional GMV-given-conversion coverage, and joint event coverage; joint coverage exceeds nominal 1-alpha by approximately 3 to 15 percentage points across alpha settings.}
  \caption{E5 outcome-layer coverage vs.\ $\alpha$ (marginal conversion, conditional GMV given conversion, and joint event). Joint empirical coverage exceeds nominal $1{-}\alpha$ across $\alpha$.}
  \label{fig:e5-y}
\end{figure}

\begin{figure}[!htbp]
  \centering
  \includegraphics[width=\linewidth]{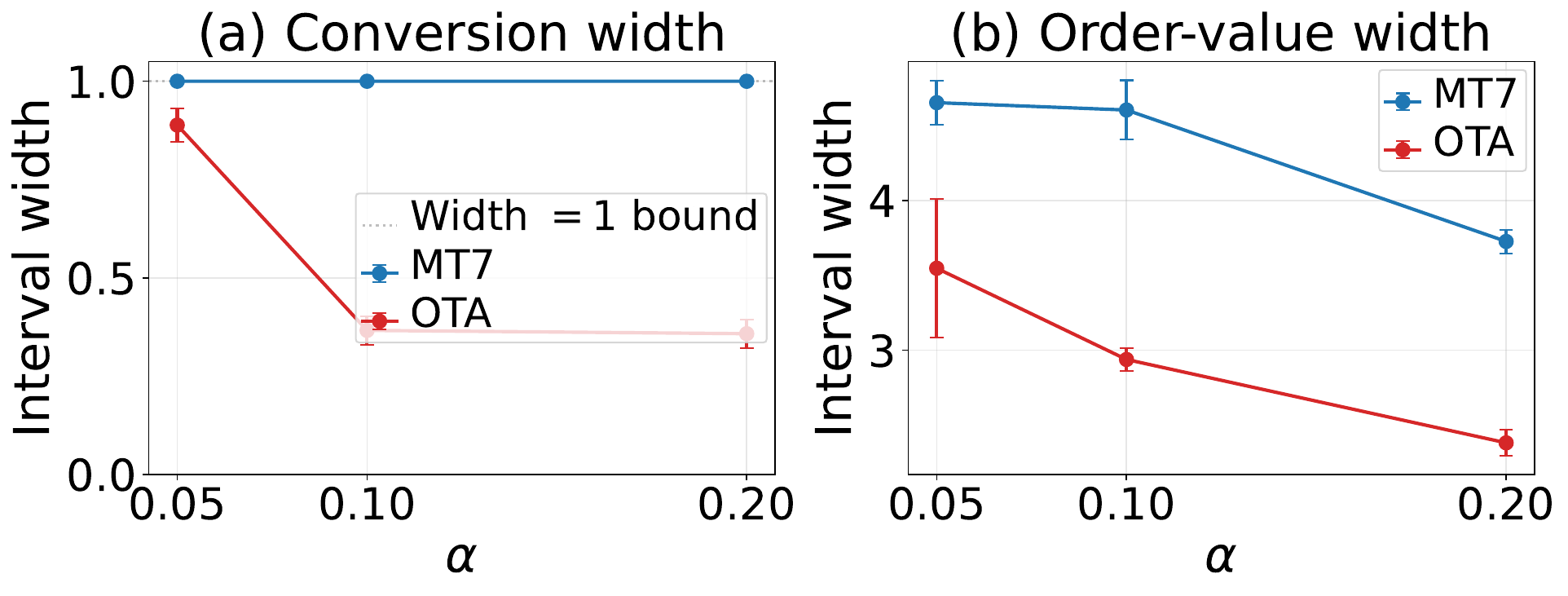}
  \Description{Line plot of mean conformal interval width versus nominal alpha for conversion and GMV heads on MT7 and OTA datasets. The conversion-head curve on OTA saturates near width one as alpha narrows toward 0.05 due to probability-axis bounding; the GMV head reported on log(1+GMV) scale narrows monotonically as alpha widens.}
  \caption{E5 mean interval width vs.\ $\alpha$: conversion head saturates near width 1 at narrow $\alpha$ on OTA; GMV head on the $\log(1{+}\mathrm{GMV})$ scale.}
  \label{fig:e5-width}
\end{figure}

Joint empirical coverage consistently \emph{exceeds} nominal $1{-}\alpha$ by 3--15 pp across $\alpha\in\{0.05,0.10,0.20\}$, reflecting conservative finite-sample CQR offsets combined with the Bonferroni union (Fig.~\ref{fig:e5-y}). On MT7 the oracle $\tau$-intervals are fully covered in these runs while $w_{\tau_g}$ remains on the order of $10^4$ currency units, so LCB-driven actions at narrow $\alpha$ are often vacuous without wider $\alpha$ or anchored point estimates (Sec.~\ref{sec:exp:e4}). On OTA, conversion-interval width on the probability scale drops sharply between $\alpha{=}0.05$ and $0.10$ (Fig.~\ref{fig:e5-width}), reflecting probability-axis saturation near width one at narrow $\alpha$.

\subsection{Computational scalability (E6)}
\label{sec:exp:e6}

\begin{figure}[!htbp]
  \centering
  \includegraphics[width=\linewidth]{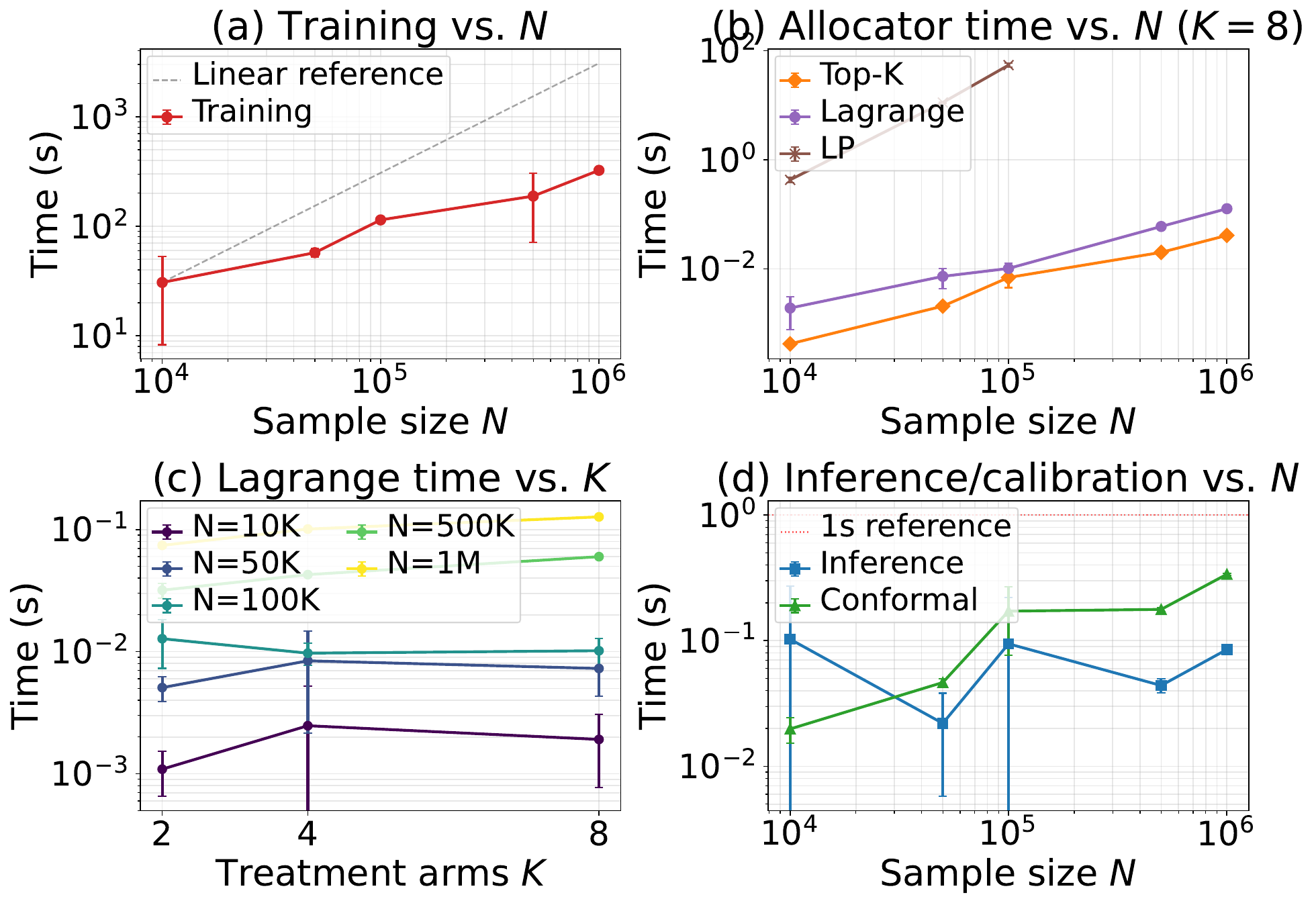}
  \Description{Four-panel log-log scaling plot of training wall-clock vs N, integer-program solvers (LP relaxation and Lagrangian dual) vs N, Lagrangian dual updates vs number of arms K, and inference plus conformal phases vs N. Training scales sublinearly between 10 thousand and 1 million users; LP fails at 500 thousand or more users due to memory while Lagrangian dual stays under 0.13 seconds at 1 million users for K equals 8.}
  \caption{E6 computational scaling (log-log): training, IP solvers, Lagrange vs.\ $K$, inference+conformal. Lagrangian allocation remains tractable at million-user scale.}
  \label{fig:e6-scaling}
\end{figure}

We measure training, conformal calibration, inference, and allocation wall-clock versus $N$ and $K$. Figure~\ref{fig:e6-scaling} plots $\log$-$\log$ scaling curves; Table~\ref{tab:e6-wallclock} excerpts $K{=}8$ timings.

\begin{table}[!htbp]
  \centering\footnotesize
  \caption{E6 wall-clock seconds ($K{=}8$, three-seed means; LP omitted where runs failed/OOM).}
  \label{tab:e6-wallclock}
  \begin{tabular}{@{}rcccc@{}}
    \toprule
    $N$ & Train & Conf.\ cal. & IP Lagrange & IP LP \\
    \midrule
    $10\mathrm{K}$  & 30.70 & 0.020 & 0.0019 & 0.429 \\
    $50\mathrm{K}$  & 57.40 & 0.047 & 0.0073 & 11.26 \\
    $500\mathrm{K}$ & 188.41 & 0.178 & 0.0599 & --- \\
    $10^{6}$       & 324.01 & 0.338 & 0.127 & --- \\
    \bottomrule
  \end{tabular}
\end{table}

Training scales sublinearly between $N{=}10^4$ and $10^6$ in our sweeps ($\approx 30\,\mathrm{s}\to 324\,\mathrm{s}$, $\sim 10\times$ wall-clock for $100\times$ users). Conformal calibration stays below one second even at $N{=}10^6$. Lagrangian dual updates stay near $0.13\,\mathrm{s}$ at one million users for $K{=}8$, whereas dense LP relaxations exceed tens of seconds already at $N{=}10^5$ and fail at larger $N$ due to memory. Production deployments therefore emphasize Lagrangian schemes with rounding while LP is reserved for moderate-scale benchmarking (including the E7 industrial headline).

\subsection{Industrial OTA: full hold-out EOM (E7)}
\label{sec:exp:e7}

We complement subsampled AUUC evidence with a \textbf{large hold-out} evaluation on de-identified Hotel-Coupon multi-arm RCT logs totaling $\approx 4.98\times 10^{6}$ exposure records from $\approx 2.79\times 10^{6}$ distinct users overall. For each of three permutation seeds we shuffle the full table, take the first $N_{\mathrm{train}}{=}50\mathrm{K}$ records for training, and retain the remaining $\approx 4.93$M exposure records per seed for evaluation. Imputation and $z$-score standardization use statistics fit \emph{only} on the training slice and applied disjointly to the hold-out so evaluation-set margins do not leak into normalization.

\paragraph{Practical operating regime.}
The data come from a de-identified hotel-coupon RCT. We treat the platform commission rate $\gamma$ as a sensitivity parameter over $[0.2,0.3]$, a band typical of online travel/coupon programs; the break-even point is $\Delta\mathrm{ROI}\!=\!1/\gamma\!\in\![3.3,5.0]$. Operating below the band ($\Delta\mathrm{ROI}\!<\!3$) means incremental commission no longer offsets subsidy cost; well above ($\Delta\mathrm{ROI}\!>\!5$) subsidies are so tight that absolute incremental GMV is rarely operationally meaningful. The $\Delta\mathrm{GMV\%}$ anchors in Table~\ref{tab:e7-ota-eom-lp} straddle this band, with smaller anchors probing the tight-budget regime where ranking quality dominates and larger anchors approaching the break-even boundary. Conclusions are insensitive to the specific $\gamma$ within $[0.2,0.3]$.

Coupon planning is expressed under alternative incremental-GMV targets $\Delta\mathrm{GMV\%}$ rather than a single universal budget. EOM evaluation follows~\cite{yan2023marketingeom}: for each dual multiplier $\alpha$, we recommend arms via the LP-relaxation KKT solution on predicted GMV lifts and costs, retain users whose randomized assignment matches the recommendation, and estimate incremental GMV with H{\'a}jek IPW on that subset relative to hold-out control mean $V_{\mathrm{ctl}}$. Let $V_\alpha$ be the H{\'a}jek-IPW mean GMV and $C_\alpha$ the corresponding H{\'a}jek-IPW mean subsidy cost on the policy-matched subset. As $\alpha$ varies, each model traces a full $(\Delta\mathrm{GMV\%},\Delta\mathrm{ROI})$ frontier, where $\Delta\mathrm{GMV\%}=100(V_\alpha-V_{\mathrm{ctl}})/V_{\mathrm{ctl}}$ and $\Delta\mathrm{ROI}=(V_\alpha-V_{\mathrm{ctl}})/C_\alpha$. The latter is the same subsidy-cost surrogate as Sec.~\ref{sec:exp:e4}, not reconciled store-level profit. Models are six multi-arm deep uplift networks---ECUP~\cite{huang2024entire}, RERUM~\cite{he2024rerum}, CFRNet~\cite{shalit2017estimating}, DragonNet~\cite{shi2019dragonnet}, EFIN~\cite{liu2023efin}, and FunnelCausalNet---trained for $30$ epochs per seed; the headline policy is \textbf{LP} allocation.

\begin{table}[!htbp]
  \centering\scriptsize
  \caption{OTA full hold-out EOM (E7): $\Delta\mathrm{ROI}$ \emph{mean$\pm$std} across three permutation seeds at representative incremental-GMV anchors read off the LP frontier ($N_{\mathrm{train}}{=}50\mathrm{K}$, $\approx 4.93\mathrm{M}$ hold-out exposure records per seed). The seeds are permutation splits of the same hold-out, not independent RCTs; per-anchor paired-bootstrap $95\%$ CIs of (FunnelCausalNet $-$ second-best) over $n{=}3$ include $0$ ($[-0.42,+0.35]$ at $10\%$ and $[-0.07,+0.38]$ at $60\%$). FunnelCausalNet has the highest seed-averaged mean at all $7/7$ anchors. A naive one-sided sign calculation gives $(1/2)^7\!\approx\!0.008$, but the anchors are correlated points on one LP frontier, so this number is descriptive rather than an independent-anchor significance test. DragonNet@$10\%$ has one seed on the frontier and no std. Boldface marks the best mean.}
  \label{tab:e7-ota-eom-lp}
  \resizebox{\columnwidth}{!}{%
  \begin{tabular}{@{}rcccccc@{}}
    \toprule
    $\Delta\mathrm{GMV\%}$ & ECUP & RERUM & CFRNet & DragonNet & EFIN & FunnelCausalNet \\
    \midrule
    $10$ & $4.13\!\pm\!0.62$ & $4.77\!\pm\!0.39$ & $4.11\!\pm\!0.08$ & $3.13\!\pm\!\mathrm{n/a}$ & $4.84\!\pm\!0.45$ & $\mathbf{4.94\!\pm\!1.08}$ \\
    $20$ & $4.00\!\pm\!0.32$ & $4.40\!\pm\!0.48$ & $3.92\!\pm\!0.05$ & $4.30\!\pm\!0.49$ & $4.23\!\pm\!0.17$ & $\mathbf{4.57\!\pm\!0.37}$ \\
    $25$ & $3.87\!\pm\!0.27$ & $4.23\!\pm\!0.39$ & $3.83\!\pm\!0.10$ & $4.12\!\pm\!0.35$ & $4.08\!\pm\!0.15$ & $\mathbf{4.44\!\pm\!0.22}$ \\
    $30$ & $3.82\!\pm\!0.24$ & $4.08\!\pm\!0.30$ & $3.75\!\pm\!0.14$ & $3.97\!\pm\!0.26$ & $3.96\!\pm\!0.14$ & $\mathbf{4.30\!\pm\!0.15}$ \\
    $40$ & $3.73\!\pm\!0.15$ & $3.81\!\pm\!0.20$ & $3.63\!\pm\!0.23$ & $3.74\!\pm\!0.18$ & $3.77\!\pm\!0.13$ & $\mathbf{4.01\!\pm\!0.17}$ \\
    $50$ & $3.60\!\pm\!0.10$ & $3.62\!\pm\!0.17$ & $3.52\!\pm\!0.17$ & $3.58\!\pm\!0.10$ & $3.62\!\pm\!0.13$ & $\mathbf{3.80\!\pm\!0.15}$ \\
    $60$ & $3.49\!\pm\!0.08$ & $3.50\!\pm\!0.11$ & $3.41\!\pm\!0.10$ & $3.49\!\pm\!0.00$ & $3.50\!\pm\!0.09$ & $\mathbf{3.60\!\pm\!0.14}$ \\
    \bottomrule
  \end{tabular}}
\end{table}

For raw-magnitude calibration, the coarse marginal per-user GMV contrast between the strongest arm and control (${\approx}92.7\%$, three-seed average) is \emph{not} the same estimand as the EOM horizontal axis $\Delta\mathrm{GMV\%}$ (an LP-policy IPW estimate at fixed dual $\alpha$). The swept LP frontiers reach different right-end extents (maximum realized $\Delta\mathrm{GMV\%}$: $\approx 72.2$ for ECUP, $74.7$ for RERUM, $86.9$ for CFRNet, $83.4$ for DragonNet, $61.8$ for EFIN, and $90.4$ for FunnelCausalNet), so curves are best read as full traces rather than single-number summaries; the larger extent under FunnelCausalNet means it can express more aggressive operating regimes that the other rankers cannot reach.

\textbf{Reading Table~\ref{tab:e7-ota-eom-lp} honestly.} FunnelCausalNet attains the highest \emph{mean} $\Delta\mathrm{ROI}$ at every anchor. The closest competitor varies by regime: at the small-anchor end ($\Delta\mathrm{GMV\%}{=}10\%$--$20\%$) it is EFIN or RERUM (both within one standard deviation), while at mid-to-large anchors ($25\%$--$60\%$) FunnelCausalNet's mean exceeds the second-best by $0.18$--$0.21$ ROI units. CFRNet sits in the lowest band at every anchor, consistent with linear-MMD balancing being designed for binary rather than tier-specific elasticities. Per-anchor paired-bootstrap CIs over three permutation seeds include $0$, so individual rows are not formally significant. The $7/7$ wins summarize the direction of one correlated frontier and do not establish cross-anchor significance. We report LP as the headline allocator because the EOM protocol~\cite{yan2023marketingeom} is defined with LP-relaxation KKT recommendations, matching both the E4 budgeted experiments and the online consistency check (Sec.~\ref{sec:exp:e8}).

\paragraph{Online consistency check.}\label{sec:exp:e8} An internal online evaluation against the platform's incumbent uplift baseline under the same LP allocator is directionally consistent with the offline EOM ordering in Table~\ref{tab:e7-ota-eom-lp}. We do not use it as headline evidence: quantitative effect sizes, per-bucket exposure ratios, and ablation traces remain unavailable under the platform agreement, so the paper's verifiable claims rely on the reported RCT/EOM aggregates and public or semi-synthetic experiments.

\subsection{Reproducibility}
\label{sec:exp:repro}

All public and semi-synthetic experiments use fixed seeds, unified configurations, and recorded run manifests; internal reruns reproduce the reported aggregates up to floating-point nondeterminism. The current version does not include a public code artifact. Public datasets remain available from their cited sources. Industrial OTA Hotel-Coupon micro-data cannot be redistributed under the platform agreement; we report aggregate metrics only, so the industrial experiment cannot be independently reproduced externally.

\section{Discussion}
\label{sec:discussion}

\paragraph{Regime-guided method choice.}
Funnel coupling is \emph{competitive} on the semi-synthetic multi-tier MT7 surface---FunnelCausalNet's AUUC\_GMV ($0.613$) is within one seed standard deviation of the leading EFIN ($0.615$)---and has the highest mean $\Delta\mathrm{ROI}$ at all $7/7$ reported industrial EOM anchors (Table~\ref{tab:e7-ota-eom-lp}). The anchors share one LP frontier and the seeds reuse one RCT through permutation splits, so this is descriptive consistency, not independent-test evidence. On single-encouragement public RCTs (e.g., Hillstrom), every tested multi-tier funnel-aware deep model underperforms revenue-centric alternatives; these results define a generalization boundary rather than evidence to assume transfer. Practical guidance is therefore regime-dependent: use hard composition when the funnel support identity is exact, consider soft penalties when logging makes that identity approximate, and evaluate revenue ranking and allocation separately across the available treatment design.

\paragraph{Ranking vs.\ calibration.}
Training emphasizes heterogeneous ordering (PEHE/AUUC); absolute ATE-style GMV calibration can remain imperfect under heavy tails. RCT-arm anchoring mitigates systematic level bias feeding budgeted objectives without retraining; forcing marginal ATE agreement would require additional regularization or doubly robust corrections and is left to future work.

\paragraph{Conformal conservatism.}
Bonferroni splits and finite-sample CQR offsets yield conservative joint coverage (empirical $1{-}\alpha$ exceeds nominal by $3$--$15$ pp in our sweeps), so narrow-$\alpha$ LCB allocation collapses toward all-control under heavy tails (Sec.~\ref{sec:exp:e4}). Wider nominal $\alpha$ or anchored point estimates remain the pragmatic pairing for actionable budgets.

\paragraph{Industrial RCTs: metric/policy interplay.}
OTA-style analyses mix subsidy costs, GMV lifts, and commission assumptions. Sec.~\ref{sec:exp:e7} adds a complementary massive hold-out EOM check: sweeping LP policies traces full $(\Delta\mathrm{GMV\%},\Delta\mathrm{ROI})$ curves, and at representative incremental-GMV anchors in $10\%$--$60\%$ FunnelCausalNet leads all multi-arm deep uplift baselines on mean $\Delta\mathrm{ROI}$ even though subsampled AUUC gaps are tight; EFIN's MT7 advantage does not transfer (its LP-frontier reach is the smallest, $\approx\!61.8\%$ vs.\ FunnelCausalNet's $\approx\!90.4\%$). Reported $\Delta\mathrm{ROI}$ is a cost-construct surrogate, not reconciled store-level profit.

\paragraph{Identification scope and limitations.}
All causal interpretations assume RCT-like randomized assignment, not observational identification. The strongest allocation evidence comes from a private industrial RCT that cannot be independently reproduced, while the public binary benchmark does not show consistent gains; the results therefore support the target multi-tier regime rather than broad dominance. Three permutation splits and correlated EOM anchors limit inferential power. Moreover, E7 uses record-level permutation splits, so repeated users can appear in both training and hold-out slices; this further limits IID and interval interpretations and motivates user-grouped splitting and clustered uncertainty in follow-up validation. The component studies isolate funnel structure and allocator variants but not a full factorial pipeline decomposition. Proposition~\ref{prop:mse-ratio} is an idealized pointwise comparison whose rate and covariance assumptions need not hold for shared neural heads. Joint conformal coverage is marginal and conservative. Finally, the observed coupon arms are discrete offers: the model does not exploit smoothness or monotonicity across a continuous coupon dose, which is an important extension when treatment intensity is not operationally tiered.

\section{Conclusion}
\label{sec:conclusion}

Coupon uplift in digital commerce must respect the funnel restriction linking conversion and GMV, cope with extreme zero inflation on revenue, and support multi-tier subsidy decisions under budgets. We presented \emph{FunnelCausalNet}, which encodes funnel composition in estimation, provides an idealized rate-gap variance comparison (Prop.~2), pairs a Lagrangian budgeted allocator with RCT-arm anchoring, and exposes a Bonferroni-union joint conformal layer plus Top-$K$ conflict screen as audit-only risk-disclosure bands. Enforcing $\mu_{\mathrm{gmv}}{=}\mu_{\mathrm{conv}}\mu_{\mathrm{val}}$ reduces GMV effect-estimation error by $18$--$48\%$ across the tested $\hat{p}\!\in\![5,45]\%$ range on Criteo-MT7 (Table~\ref{tab:e2b-prop2}); on industrial multi-arm RCT logs, FunnelCausalNet has the highest seed-averaged mean LP-frontier $\Delta\mathrm{ROI}$ at all $7/7$ reported anchors, although their correlation and the three permutation splits preclude an independent-anchor significance claim. FunnelCausalNet does not lead every semi-synthetic or public benchmark, so the evidence supports a practically important multi-tier regime rather than universal superiority. Future work includes continuous-dose extensions, doubly robust calibration under sparse converters, and broader public or online validation.

\appendix
\section{Proof sketch for Proposition~2 (delta method under a rate gap)}
\label{app:proof}
Fix $(X,T){=}(x,t)$ and let $\mu_g{=}\mathbb{E}[Y^{g}\mid X{=}x,T{=}t]{=}p\mu_v$. Under (A1)--(A2), $\mu_g$ is identified from the RCT logs and the population funnel composition follows by iterated expectations. Prop.~1 separates the within-converter and Bernoulli switching terms. Under (A4), the direct estimator has pointwise asymptotic variance
\begin{equation*}
  \mathrm{Var}(\hat\mu_g^{\mathrm{direct}})
  \sim \{p\sigma_v^2+p(1-p)\mu_v^2\}/r_n,
\end{equation*}
with negligible bias under standard undersmoothing. The delta method gives
\begin{align*}
  \mathrm{Var}(\hat\mu_g^{\mathrm{funnel}})
  ={}&\mu_v^2\mathrm{Var}(\hat p)
  +p^2\mathrm{Var}(\hat\mu_v)\\
  &+2p\mu_v\mathrm{Cov}(\hat p,\hat\mu_v),
\end{align*}
where $\mathrm{Var}(\hat p){=}O(n^{-1})$ under (A3) and $\mathrm{Var}(\hat\mu_v){=}\sigma_v^2/(pr_n)$ under (A4). Assumption (A5) sets the covariance to zero for estimates fit on independent folds; otherwise it requires $o(1/r_n)$. Cauchy--Schwarz gives the upper order $O((nr_n)^{-1/2})$, which is $o(1/r_n)$ when $r_n{=}o(n)$. The conv-head variance contribution is also $o(1/r_n)$, so the funnel variance reduces to $p\sigma_v^2/r_n+o(1/r_n)$ and the leading-order ratio gives Eq.~\eqref{eq:mse-ratio}.

This calculation assumes negligible bias. Correlated systematic errors from shared neural representations can change the finite-sample product error and are not covered by the proposition. The argument follows the standard hurdle/two-part decomposition pattern of~\cite{cragg1971some,mullahy1986specification,lambert1992zero}; here it is a regime heuristic for coupon uplift rather than a neural-model guarantee.

\FloatBarrier
\section*{GenAI Usage Disclosure}
In accordance with the CIKM 2026 generative-AI policy, we disclose that generative AI assistants (large language models) were used during manuscript preparation for (i) language polishing, consistency checks, and editorial restructuring of author-provided technical text, including clarification of the scope and assumptions of theoretical and empirical claims; and (ii) refactoring of plotting and CSV-aggregation utilities. The authors determined the technical contributions, derivations, experimental designs, analyses, and result interpretations; reviewed and revised all generated text and code before inclusion; and take full responsibility for the manuscript. No GenAI-generated output was used as data or experimental evidence.

\bibliographystyle{ACM-Reference-Format}
\bibliography{refs}


\begin{thebibliography}{44}


\ifx \showCODEN    \undefined \def \showCODEN     #1{\unskip}     \fi
\ifx \showISBNx    \undefined \def \showISBNx     #1{\unskip}     \fi
\ifx \showISBNxiii \undefined \def \showISBNxiii  #1{\unskip}     \fi
\ifx \showISSN     \undefined \def \showISSN      #1{\unskip}     \fi
\ifx \showLCCN     \undefined \def \showLCCN      #1{\unskip}     \fi
\ifx \shownote     \undefined \def \shownote      #1{#1}          \fi
\ifx \showarticletitle \undefined \def \showarticletitle #1{#1}   \fi
\ifx \showURL      \undefined \def \showURL       {\relax}        \fi
\providecommand\bibfield[2]{#2}
\providecommand\bibinfo[2]{#2}
\providecommand\natexlab[1]{#1}
\providecommand\showeprint[2][]{arXiv:#2}

\bibitem[Alaa and van~der Schaar(2019)]%
        {alaa2019validation}
\bibfield{author}{\bibinfo{person}{Ahmed~M. Alaa} {and}
  \bibinfo{person}{Mihaela van~der Schaar}.} \bibinfo{year}{2019}\natexlab{}.
\newblock \showarticletitle{Validating causal inference models via conformal
  prediction}.
\newblock \bibinfo{journal}{\emph{Proceedings of Machine Learning Research}}
  \bibinfo{volume}{89} (\bibinfo{year}{2019}), \bibinfo{pages}{124--133}.
\newblock


\bibitem[Albert and Goldenberg(2022)]%
        {albert2022ecommerce}
\bibfield{author}{\bibinfo{person}{Javier Albert} {and} \bibinfo{person}{Dmitri
  Goldenberg}.} \bibinfo{year}{2022}\natexlab{}.
\newblock \showarticletitle{E-commerce promotions personalization via online
  multiple-choice knapsack with uplift modeling}. In
  \bibinfo{booktitle}{\emph{Proceedings of the 31st ACM International
  Conference on Information and Knowledge Management}}.
  \bibinfo{pages}{2864--2872}.
\newblock
\href{https://doi.org/10.1145/3511808.3557229}{doi:\nolinkurl{10.1145/3511808.3557229}}


\bibitem[{Alibaba Tianchi}(2018)]%
        {tianchio2o2018}
\bibfield{author}{\bibinfo{person}{{Alibaba Tianchi}}.}
  \bibinfo{year}{2018}\natexlab{}.
\newblock \bibinfo{title}{{Tianchi} {O2O} Coupon Usage Forecast}.
\newblock \bibinfo{howpublished}{Public competition dataset,
  \url{https://tianchi.aliyun.com/competition/entrance/231593}; observational
  logs with coupon-redemption labels and no continuous GMV outcome}.
\newblock


\bibitem[Athey et~al\mbox{.}(2019)]%
        {athey2019generalized}
\bibfield{author}{\bibinfo{person}{Susan Athey}, \bibinfo{person}{Julie
  Tibshirani}, {and} \bibinfo{person}{Stefan Wager}.}
  \bibinfo{year}{2019}\natexlab{}.
\newblock \showarticletitle{Generalized random forests}.
\newblock \bibinfo{journal}{\emph{The Annals of Statistics}}
  \bibinfo{volume}{47}, \bibinfo{number}{2} (\bibinfo{year}{2019}),
  \bibinfo{pages}{1148--1178}.
\newblock
\href{https://doi.org/10.1214/18-AOS1709}{doi:\nolinkurl{10.1214/18-AOS1709}}


\bibitem[Betlei et~al\mbox{.}(2021)]%
        {betlei2021uplift}
\bibfield{author}{\bibinfo{person}{Artem Betlei}, \bibinfo{person}{Eustache
  Diemert}, {and} \bibinfo{person}{Massih-Reza Amini}.}
  \bibinfo{year}{2021}\natexlab{}.
\newblock \showarticletitle{Uplift Modeling with Generalization Guarantees}. In
  \bibinfo{booktitle}{\emph{Proceedings of the 27th ACM SIGKDD Conference on
  Knowledge Discovery \& Data Mining}}. \bibinfo{pages}{55--65}.
\newblock
\href{https://doi.org/10.1145/3447548.3467395}{doi:\nolinkurl{10.1145/3447548.3467395}}


\bibitem[Chen et~al\mbox{.}(2024)]%
        {chen2024upliftrec}
\bibfield{author}{\bibinfo{person}{Jiaju Chen}, \bibinfo{person}{Wenjie Wang},
  \bibinfo{person}{Chongming Gao}, \bibinfo{person}{Peng Wu},
  \bibinfo{person}{Jianxiong Wei}, {and} \bibinfo{person}{Qingsong Hua}.}
  \bibinfo{year}{2024}\natexlab{}.
\newblock \showarticletitle{Treatment Effect Estimation for User Interest
  Exploration on Recommender Systems}. In \bibinfo{booktitle}{\emph{Proceedings
  of the 47th International ACM SIGIR Conference on Research and Development in
  Information Retrieval}}. \bibinfo{pages}{1861--1871}.
\newblock
\href{https://doi.org/10.1145/3626772.3657736}{doi:\nolinkurl{10.1145/3626772.3657736}}


\bibitem[Chernozhukov et~al\mbox{.}(2018)]%
        {chernozhukov2018double}
\bibfield{author}{\bibinfo{person}{Victor Chernozhukov}, \bibinfo{person}{Denis
  Chetverikov}, \bibinfo{person}{Mert Demirer}, \bibinfo{person}{Esther Duflo},
  \bibinfo{person}{Christian Hansen}, \bibinfo{person}{Whitney Newey}, {and}
  \bibinfo{person}{James Robins}.} \bibinfo{year}{2018}\natexlab{}.
\newblock \showarticletitle{Double/debiased machine learning for treatment and
  structural parameters}.
\newblock \bibinfo{journal}{\emph{The Econometrics Journal}}
  \bibinfo{volume}{21}, \bibinfo{number}{1} (\bibinfo{year}{2018}),
  \bibinfo{pages}{C1--C68}.
\newblock
\href{https://doi.org/10.1111/ectj.12097}{doi:\nolinkurl{10.1111/ectj.12097}}


\bibitem[Cragg(1971)]%
        {cragg1971some}
\bibfield{author}{\bibinfo{person}{John~G. Cragg}.}
  \bibinfo{year}{1971}\natexlab{}.
\newblock \showarticletitle{Some statistical models for limited dependent
  variables with application to the demand for durable goods}.
\newblock \bibinfo{journal}{\emph{Econometrica}} \bibinfo{volume}{39},
  \bibinfo{number}{5} (\bibinfo{year}{1971}), \bibinfo{pages}{829--844}.
\newblock
\href{https://doi.org/10.2307/1909582}{doi:\nolinkurl{10.2307/1909582}}


\bibitem[Devriendt et~al\mbox{.}(2018)]%
        {devriendt2018literature}
\bibfield{author}{\bibinfo{person}{Floris Devriendt}, \bibinfo{person}{Darie
  Moldovan}, {and} \bibinfo{person}{Wouter Verbeke}.}
  \bibinfo{year}{2018}\natexlab{}.
\newblock \showarticletitle{A literature survey and experimental evaluation of
  the state-of-the-art in uplift modeling: A stepping stone toward the
  development of prescriptive analytics}.
\newblock \bibinfo{journal}{\emph{Big Data}} \bibinfo{volume}{6},
  \bibinfo{number}{1} (\bibinfo{year}{2018}), \bibinfo{pages}{13--41}.
\newblock
\href{https://doi.org/10.1089/big.2017.0104}{doi:\nolinkurl{10.1089/big.2017.0104}}


\bibitem[Diemert et~al\mbox{.}(2018)]%
        {diemert2018large}
\bibfield{author}{\bibinfo{person}{Eustache Diemert}, \bibinfo{person}{Artem
  Betlei}, \bibinfo{person}{Christophe Renaudin}, {and}
  \bibinfo{person}{Massih-Reza Amini}.} \bibinfo{year}{2018}\natexlab{}.
\newblock \showarticletitle{A large scale benchmark for uplift modeling}. In
  \bibinfo{booktitle}{\emph{Proceedings of the AdKDD and TargetAd Workshop, KDD
  '18}}. \bibinfo{publisher}{ACM}, \bibinfo{address}{London, United Kingdom},
  \bibinfo{pages}{1--6}.
\newblock


\bibitem[Gutierrez and G{\'e}rardy(2017)]%
        {gutierrez2017causal}
\bibfield{author}{\bibinfo{person}{Pierre Gutierrez} {and}
  \bibinfo{person}{Jean-Yves G{\'e}rardy}.} \bibinfo{year}{2017}\natexlab{}.
\newblock \showarticletitle{Causal inference and uplift modelling: A review of
  the literature}. In \bibinfo{booktitle}{\emph{Proceedings of the
  International Conference on Predictive Applications and APIs (PAPIs)}}
  \emph{(\bibinfo{series}{Proceedings of Machine Learning Research},
  Vol.~\bibinfo{volume}{67})}. \bibinfo{publisher}{PMLR},
  \bibinfo{pages}{1--13}.
\newblock


\bibitem[He et~al\mbox{.}(2024)]%
        {he2024rerum}
\bibfield{author}{\bibinfo{person}{Bowei He}, \bibinfo{person}{Yunpeng Weng},
  \bibinfo{person}{Xing Tang}, \bibinfo{person}{Ziqiang Cui},
  \bibinfo{person}{Zexu Sun}, \bibinfo{person}{Liang Chen},
  \bibinfo{person}{Xiuqiang He}, {and} \bibinfo{person}{Chen Ma}.}
  \bibinfo{year}{2024}\natexlab{}.
\newblock \showarticletitle{Rankability-enhanced Revenue Uplift Modeling
  Framework for Online Marketing}. In \bibinfo{booktitle}{\emph{Proceedings of
  the 30th ACM SIGKDD Conference on Knowledge Discovery and Data Mining}}.
  \bibinfo{pages}{5093--5104}.
\newblock
\href{https://doi.org/10.1145/3637528.3671516}{doi:\nolinkurl{10.1145/3637528.3671516}}


\bibitem[Hill(2011)]%
        {hill2011bayesian}
\bibfield{author}{\bibinfo{person}{Jennifer~L. Hill}.}
  \bibinfo{year}{2011}\natexlab{}.
\newblock \showarticletitle{Bayesian nonparametric modeling for causal
  inference}.
\newblock \bibinfo{journal}{\emph{Journal of Computational and Graphical
  Statistics}} \bibinfo{volume}{20}, \bibinfo{number}{1}
  (\bibinfo{year}{2011}), \bibinfo{pages}{217--240}.
\newblock
\href{https://doi.org/10.1198/jcgs.2010.08162}{doi:\nolinkurl{10.1198/jcgs.2010.08162}}


\bibitem[Hillstrom(2008)]%
        {hillstrom2008minethatdata}
\bibfield{author}{\bibinfo{person}{Kevin Hillstrom}.}
  \bibinfo{year}{2008}\natexlab{}.
\newblock \bibinfo{title}{The {MineThatData} {E}-Mail Analytics And Data Mining
  Challenge}.
\newblock \bibinfo{howpublished}{Online dataset and blog post}.
\newblock
\newblock
\shownote{\url{https://blog.minethatdata.com/2008/03/minethatdata-e-mail-analytics-and-data.html};
  64\,000-customer randomized email-marketing dataset widely used as an
  uplift-modeling benchmark}.


\bibitem[Huang et~al\mbox{.}(2024)]%
        {huang2024entire}
\bibfield{author}{\bibinfo{person}{Yinqiu Huang}, \bibinfo{person}{Shuli Wang},
  \bibinfo{person}{Min Gao}, \bibinfo{person}{Xue Wei},
  \bibinfo{person}{Changhao Li}, \bibinfo{person}{Chuan Luo},
  \bibinfo{person}{Yinhua Zhu}, \bibinfo{person}{Xiong Xiao}, {and}
  \bibinfo{person}{Yi Luo}.} \bibinfo{year}{2024}\natexlab{}.
\newblock \showarticletitle{Entire Chain Uplift Modeling with Context-Enhanced
  Learning for Intelligent Marketing}. In \bibinfo{booktitle}{\emph{Companion
  Proceedings of the ACM Web Conference 2024}}. \bibinfo{pages}{226--234}.
\newblock
\href{https://doi.org/10.1145/3589335.3648320}{doi:\nolinkurl{10.1145/3589335.3648320}}


\bibitem[Kong et~al\mbox{.}(2026)]%
        {kong2026saco}
\bibfield{author}{\bibinfo{person}{Li Kong}, \bibinfo{person}{Bingzhe Wang},
  \bibinfo{person}{Zhou Chen}, \bibinfo{person}{Suhan Hu},
  \bibinfo{person}{Yuchao Ma}, \bibinfo{person}{Qi Qi},
  \bibinfo{person}{Suoyuan Song}, {and} \bibinfo{person}{Bicheng Jin}.}
  \bibinfo{year}{2026}\natexlab{}.
\newblock \showarticletitle{{SACO}: Sequence-Aware Constrained Optimization
  Framework for Coupon Distribution in {E}-commerce}. In
  \bibinfo{booktitle}{\emph{Proceedings of the AAAI Conference on Artificial
  Intelligence}}, Vol.~\bibinfo{volume}{40}. \bibinfo{pages}{15027--15035}.
\newblock
\href{https://doi.org/10.1609/aaai.v40i17.38525}{doi:\nolinkurl{10.1609/aaai.v40i17.38525}}


\bibitem[K{\"u}nzel et~al\mbox{.}(2019)]%
        {kunzel2019metalearners}
\bibfield{author}{\bibinfo{person}{S{\"o}ren~R. K{\"u}nzel},
  \bibinfo{person}{Jasjeet~S. Sekhon}, \bibinfo{person}{Peter~J. Bickel}, {and}
  \bibinfo{person}{Bin Yu}.} \bibinfo{year}{2019}\natexlab{}.
\newblock \showarticletitle{Metalearners for estimating heterogeneous treatment
  effects using machine learning}.
\newblock \bibinfo{journal}{\emph{Proceedings of the National Academy of
  Sciences}} \bibinfo{volume}{116}, \bibinfo{number}{10}
  (\bibinfo{year}{2019}), \bibinfo{pages}{4156--4165}.
\newblock
\href{https://doi.org/10.1073/pnas.1804597116}{doi:\nolinkurl{10.1073/pnas.1804597116}}


\bibitem[Lambert(1992)]%
        {lambert1992zero}
\bibfield{author}{\bibinfo{person}{Diane Lambert}.}
  \bibinfo{year}{1992}\natexlab{}.
\newblock \showarticletitle{Zero-inflated {Poisson} regression, with an
  application to defects in manufacturing}.
\newblock \bibinfo{journal}{\emph{Technometrics}} \bibinfo{volume}{34},
  \bibinfo{number}{1} (\bibinfo{year}{1992}), \bibinfo{pages}{1--14}.
\newblock
\href{https://doi.org/10.2307/1269547}{doi:\nolinkurl{10.2307/1269547}}


\bibitem[Lei and Cand{\`e}s(2021)]%
        {lei2021conformal}
\bibfield{author}{\bibinfo{person}{Lihua Lei} {and}
  \bibinfo{person}{Emmanuel~J. Cand{\`e}s}.} \bibinfo{year}{2021}\natexlab{}.
\newblock \showarticletitle{Conformal inference of counterfactuals and
  individual treatment effects}.
\newblock \bibinfo{journal}{\emph{Journal of the Royal Statistical Society:
  Series B (Statistical Methodology)}} \bibinfo{volume}{83},
  \bibinfo{number}{5} (\bibinfo{year}{2021}), \bibinfo{pages}{911--938}.
\newblock
\href{https://doi.org/10.1111/rssb.12445}{doi:\nolinkurl{10.1111/rssb.12445}}


\bibitem[{Lenta Group}(2020)]%
        {lentauplift2020}
\bibfield{author}{\bibinfo{person}{{Lenta Group}}.}
  \bibinfo{year}{2020}\natexlab{}.
\newblock \bibinfo{title}{Lenta uplift modeling dataset}.
\newblock \bibinfo{howpublished}{Public retail-loyalty dataset,
  \url{https://github.com/maks-sh/scikit-uplift}; binary SMS encouragement, no
  continuous-spend supervision}.
\newblock


\bibitem[Li et~al\mbox{.}(2020)]%
        {li2020spending}
\bibfield{author}{\bibinfo{person}{Liangwei Li}, \bibinfo{person}{Liucheng
  Sun}, \bibinfo{person}{Chenwei Weng}, \bibinfo{person}{Chengfu Huo}, {and}
  \bibinfo{person}{Weijun Ren}.} \bibinfo{year}{2020}\natexlab{}.
\newblock \showarticletitle{Spending Money Wisely: Online Electronic Coupon
  Allocation based on Real-Time User Intent Detection}. In
  \bibinfo{booktitle}{\emph{Proceedings of the 29th ACM International
  Conference on Information and Knowledge Management}}.
  \bibinfo{pages}{2597--2604}.
\newblock
\href{https://doi.org/10.1145/3340531.3412745}{doi:\nolinkurl{10.1145/3340531.3412745}}


\bibitem[Liu et~al\mbox{.}(2023)]%
        {liu2023efin}
\bibfield{author}{\bibinfo{person}{Dugang Liu}, \bibinfo{person}{Xing Tang},
  \bibinfo{person}{Han Gao}, \bibinfo{person}{Fuyuan Lyu}, {and}
  \bibinfo{person}{Xiuqiang He}.} \bibinfo{year}{2023}\natexlab{}.
\newblock \showarticletitle{Explicit Feature Interaction-aware Uplift Network
  for Online Marketing}. In \bibinfo{booktitle}{\emph{Proceedings of the 29th
  ACM SIGKDD Conference on Knowledge Discovery and Data Mining}}.
  \bibinfo{pages}{4507--4515}.
\newblock
\href{https://doi.org/10.1145/3580305.3599820}{doi:\nolinkurl{10.1145/3580305.3599820}}


\bibitem[Ma et~al\mbox{.}(2018)]%
        {ma2018esmm}
\bibfield{author}{\bibinfo{person}{Xiao Ma}, \bibinfo{person}{Liqin Zhao},
  \bibinfo{person}{Guan Huang}, \bibinfo{person}{Zhi Wang},
  \bibinfo{person}{Zelin Hu}, \bibinfo{person}{Xiaoqiang Zhu}, {and}
  \bibinfo{person}{Kun Gai}.} \bibinfo{year}{2018}\natexlab{}.
\newblock \showarticletitle{Entire Space Multi-Task Model: An Effective
  Approach for Estimating Post-Click Conversion Rate}. In
  \bibinfo{booktitle}{\emph{Proceedings of the 41st International ACM SIGIR
  Conference on Research and Development in Information Retrieval}}.
  \bibinfo{pages}{1137--1140}.
\newblock
\href{https://doi.org/10.1145/3209978.3210104}{doi:\nolinkurl{10.1145/3209978.3210104}}


\bibitem[{MegaFon}(2021)]%
        {megafonuplift2021}
\bibfield{author}{\bibinfo{person}{{MegaFon}}.}
  \bibinfo{year}{2021}\natexlab{}.
\newblock \bibinfo{title}{{MegaFon} Uplift Competition dataset}.
\newblock \bibinfo{howpublished}{Public synthetic uplift challenge dataset;
  binary treatment, no tier-strength axis}.
\newblock


\bibitem[Mondal et~al\mbox{.}(2022)]%
        {mondal2022memento}
\bibfield{author}{\bibinfo{person}{Abhirup Mondal}, \bibinfo{person}{Anirban
  Majumder}, {and} \bibinfo{person}{Vineet Chaoji}.}
  \bibinfo{year}{2022}\natexlab{}.
\newblock \showarticletitle{{MEMENTO}: Neural Model for Estimating Individual
  Treatment Effects for Multiple Treatments}. In
  \bibinfo{booktitle}{\emph{Proceedings of the 31st ACM International
  Conference on Information \& Knowledge Management}}.
  \bibinfo{pages}{3381--3390}.
\newblock
\href{https://doi.org/10.1145/3511808.3557125}{doi:\nolinkurl{10.1145/3511808.3557125}}


\bibitem[Mullahy(1986)]%
        {mullahy1986specification}
\bibfield{author}{\bibinfo{person}{John Mullahy}.}
  \bibinfo{year}{1986}\natexlab{}.
\newblock \showarticletitle{Specification and testing of some modified count
  data models}.
\newblock \bibinfo{journal}{\emph{Journal of Econometrics}}
  \bibinfo{volume}{33}, \bibinfo{number}{3} (\bibinfo{year}{1986}),
  \bibinfo{pages}{341--365}.
\newblock
\href{https://doi.org/10.1016/0304-4076(86)90002-3}{doi:\nolinkurl{10.1016/0304-4076(86)90002-3}}


\bibitem[Nie and Wager(2021)]%
        {nie2021quasi}
\bibfield{author}{\bibinfo{person}{Xinkun Nie} {and} \bibinfo{person}{Stefan
  Wager}.} \bibinfo{year}{2021}\natexlab{}.
\newblock \showarticletitle{Quasi-oracle estimation of heterogeneous treatment
  effects}.
\newblock \bibinfo{journal}{\emph{Biometrika}} \bibinfo{volume}{108},
  \bibinfo{number}{2} (\bibinfo{year}{2021}), \bibinfo{pages}{299--319}.
\newblock
\href{https://doi.org/10.1093/biomet/asaa076}{doi:\nolinkurl{10.1093/biomet/asaa076}}


\bibitem[Romano et~al\mbox{.}(2019)]%
        {roman2019conformalized}
\bibfield{author}{\bibinfo{person}{Yaniv Romano}, \bibinfo{person}{Evan
  Patterson}, {and} \bibinfo{person}{Emmanuel Cand{\`e}s}.}
  \bibinfo{year}{2019}\natexlab{}.
\newblock \showarticletitle{Conformalized quantile regression}. In
  \bibinfo{booktitle}{\emph{Advances in Neural Information Processing
  Systems}}, Vol.~\bibinfo{volume}{32}. \bibinfo{pages}{3543--3553}.
\newblock


\bibitem[Saito et~al\mbox{.}(2021)]%
        {saito2021open}
\bibfield{author}{\bibinfo{person}{Yuta Saito}, \bibinfo{person}{Shunsuke
  Aihara}, \bibinfo{person}{Megumi Matsutani}, {and} \bibinfo{person}{Yusuke
  Narita}.} \bibinfo{year}{2021}\natexlab{}.
\newblock \showarticletitle{Open Bandit Dataset and Pipeline: Towards realistic
  and reproducible off-policy evaluation}. In
  \bibinfo{booktitle}{\emph{Advances in Neural Information Processing Systems
  Datasets and Benchmarks Track}}.
\newblock


\bibitem[Shalit et~al\mbox{.}(2017)]%
        {shalit2017estimating}
\bibfield{author}{\bibinfo{person}{Uri Shalit}, \bibinfo{person}{Fredrik~D.
  Johansson}, {and} \bibinfo{person}{David Sontag}.}
  \bibinfo{year}{2017}\natexlab{}.
\newblock \showarticletitle{Estimating individual treatment effect:
  Generalization bounds and algorithms}. In
  \bibinfo{booktitle}{\emph{Proceedings of the 34th International Conference on
  Machine Learning}}. \bibinfo{publisher}{PMLR}, \bibinfo{pages}{3076--3085}.
\newblock


\bibitem[Shi et~al\mbox{.}(2019)]%
        {shi2019dragonnet}
\bibfield{author}{\bibinfo{person}{Claudia Shi}, \bibinfo{person}{David~M.
  Blei}, {and} \bibinfo{person}{Victor Veitch}.}
  \bibinfo{year}{2019}\natexlab{}.
\newblock \showarticletitle{Adapting Neural Networks for the Estimation of
  Treatment Effects}. In \bibinfo{booktitle}{\emph{Advances in Neural
  Information Processing Systems}}, Vol.~\bibinfo{volume}{32}.
\newblock


\bibitem[Sun et~al\mbox{.}(2024)]%
        {sun2024e3ir}
\bibfield{author}{\bibinfo{person}{Zexu Sun}, \bibinfo{person}{Hao Yang},
  \bibinfo{person}{Dugang Liu}, \bibinfo{person}{Yunpeng Weng},
  \bibinfo{person}{Xing Tang}, {and} \bibinfo{person}{Xiuqiang He}.}
  \bibinfo{year}{2024}\natexlab{}.
\newblock \showarticletitle{End-to-End Cost-Effective Incentive Recommendation
  under Budget Constraint with Uplift Modeling}. In
  \bibinfo{booktitle}{\emph{Proceedings of the 18th ACM Conference on
  Recommender Systems (RecSys '24)}}. \bibinfo{pages}{560--569}.
\newblock
\href{https://doi.org/10.1145/3640457.3688147}{doi:\nolinkurl{10.1145/3640457.3688147}}


\bibitem[Tu et~al\mbox{.}(2024)]%
        {tu2024realtime}
\bibfield{author}{\bibinfo{person}{Jinglong Tu}, \bibinfo{person}{Qi Qi},
  \bibinfo{person}{Zhilin Li}, {and} \bibinfo{person}{Shuanglong Fan}.}
  \bibinfo{year}{2024}\natexlab{}.
\newblock \bibinfo{title}{Data-driven real-time coupon allocation in the online
  platform}.
\newblock
\showeprint[arxiv]{2406.05987}~[cs.LG]


\bibitem[Vovk et~al\mbox{.}(2005)]%
        {vovk2005algorithmic}
\bibfield{author}{\bibinfo{person}{Vladimir Vovk}, \bibinfo{person}{Alexander
  Gammerman}, {and} \bibinfo{person}{Glenn Shafer}.}
  \bibinfo{year}{2005}\natexlab{}.
\newblock \bibinfo{booktitle}{\emph{Algorithmic Learning in a Random World}}.
\newblock \bibinfo{publisher}{Springer}, \bibinfo{address}{New York}.
\newblock


\bibitem[Wager and Athey(2018)]%
        {wager2018estimation}
\bibfield{author}{\bibinfo{person}{Stefan Wager} {and} \bibinfo{person}{Susan
  Athey}.} \bibinfo{year}{2018}\natexlab{}.
\newblock \showarticletitle{Estimation and inference of heterogeneous treatment
  effects using random forests}.
\newblock \bibinfo{journal}{\emph{J. Amer. Statist. Assoc.}}
  \bibinfo{volume}{113}, \bibinfo{number}{523} (\bibinfo{year}{2018}),
  \bibinfo{pages}{1228--1242}.
\newblock
\href{https://doi.org/10.1080/01621459.2017.1319839}{doi:\nolinkurl{10.1080/01621459.2017.1319839}}


\bibitem[Wang et~al\mbox{.}(2022)]%
        {wang2022escm2}
\bibfield{author}{\bibinfo{person}{Hao Wang}, \bibinfo{person}{Tai-Wei Chang},
  \bibinfo{person}{Tianqiao Liu}, \bibinfo{person}{Jianmin Huang},
  \bibinfo{person}{Zhichao Chen}, \bibinfo{person}{Chao Yu},
  \bibinfo{person}{Ruopeng Li}, {and} \bibinfo{person}{Wei Chu}.}
  \bibinfo{year}{2022}\natexlab{}.
\newblock \showarticletitle{{ESCM2}: Entire Space Counterfactual Multi-Task
  Model for Post-Click Conversion Rate Estimation}. In
  \bibinfo{booktitle}{\emph{Proceedings of the 45th International ACM SIGIR
  Conference on Research and Development in Information Retrieval}}.
  \bibinfo{pages}{363--372}.
\newblock
\href{https://doi.org/10.1145/3477495.3531972}{doi:\nolinkurl{10.1145/3477495.3531972}}


\bibitem[Wei et~al\mbox{.}(2024)]%
        {wei2024mtmt}
\bibfield{author}{\bibinfo{person}{Yuxiang Wei}, \bibinfo{person}{Zhaoxin Qiu},
  \bibinfo{person}{Yingjie Li}, \bibinfo{person}{Yuke Sun}, {and}
  \bibinfo{person}{Xiaoling Li}.} \bibinfo{year}{2024}\natexlab{}.
\newblock \bibinfo{title}{Multi-Treatment Multi-Task Uplift Modeling for
  Enhancing User Growth}.
\newblock
\showeprint[arxiv]{2408.12803}~[cs.LG]


\bibitem[Weng et~al\mbox{.}(2024)]%
        {weng2024optdist}
\bibfield{author}{\bibinfo{person}{Yunpeng Weng}, \bibinfo{person}{Xing Tang},
  \bibinfo{person}{Zhenhao Xu}, \bibinfo{person}{Fuyuan Lyu},
  \bibinfo{person}{Dugang Liu}, \bibinfo{person}{Zexu Sun}, {and}
  \bibinfo{person}{Xiuqiang He}.} \bibinfo{year}{2024}\natexlab{}.
\newblock \showarticletitle{{OptDist}: Learning Optimal Distribution for
  Customer Lifetime Value Prediction}. In \bibinfo{booktitle}{\emph{Proceedings
  of the 33rd ACM International Conference on Information and Knowledge
  Management}}.
\newblock
\href{https://doi.org/10.1145/3627673.3679712}{doi:\nolinkurl{10.1145/3627673.3679712}}


\bibitem[Yan et~al\mbox{.}(2023)]%
        {yan2023marketingeom}
\bibfield{author}{\bibinfo{person}{Ziang Yan}, \bibinfo{person}{Shusen Wang},
  \bibinfo{person}{Guorui Zhou}, \bibinfo{person}{Jingjian Lin}, {and}
  \bibinfo{person}{Peng Jiang}.} \bibinfo{year}{2023}\natexlab{}.
\newblock \bibinfo{title}{An End-to-End Framework for Marketing Effectiveness
  Optimization under Budget Constraint}.
\newblock
\showeprint[arxiv]{2302.04477}~[cs.LG]


\bibitem[Yao et~al\mbox{.}(2024)]%
        {yao2024cio}
\bibfield{author}{\bibinfo{person}{Dong Yao}, \bibinfo{person}{Caizhi Tang},
  \bibinfo{person}{Qing Cui}, {and} \bibinfo{person}{Longfei Li}.}
  \bibinfo{year}{2024}\natexlab{}.
\newblock \showarticletitle{Combining Incomplete Observational and Randomized
  Data for Heterogeneous Treatment Effects}. In
  \bibinfo{booktitle}{\emph{Proceedings of the 33rd ACM International
  Conference on Information and Knowledge Management}}.
\newblock
\href{https://doi.org/10.1145/3627673.3679593}{doi:\nolinkurl{10.1145/3627673.3679593}}


\bibitem[Zhang et~al\mbox{.}(2021)]%
        {zhang2021unified}
\bibfield{author}{\bibinfo{person}{Weijia Zhang}, \bibinfo{person}{Jiuyong Li},
  {and} \bibinfo{person}{Lin Liu}.} \bibinfo{year}{2021}\natexlab{}.
\newblock \showarticletitle{A unified survey of treatment effect heterogeneity
  modelling and uplift modelling}.
\newblock \bibinfo{journal}{\emph{Comput. Surveys}} \bibinfo{volume}{54},
  \bibinfo{number}{8} (\bibinfo{year}{2021}), \bibinfo{pages}{1--36}.
\newblock
\href{https://doi.org/10.1145/3466818}{doi:\nolinkurl{10.1145/3466818}}


\bibitem[Zhao et~al\mbox{.}(2019)]%
        {zhao2019unified}
\bibfield{author}{\bibinfo{person}{Kui Zhao}, \bibinfo{person}{Junhao Wang},
  \bibinfo{person}{Bo Long}, \bibinfo{person}{Jian Xu}, {and}
  \bibinfo{person}{Kun Gai}.} \bibinfo{year}{2019}\natexlab{}.
\newblock \showarticletitle{A unified framework for marketing budget
  allocation}. In \bibinfo{booktitle}{\emph{Proceedings of the 25th ACM SIGKDD
  International Conference on Knowledge Discovery \& Data Mining}}.
  \bibinfo{pages}{2820--2828}.
\newblock
\href{https://doi.org/10.1145/3292500.3330787}{doi:\nolinkurl{10.1145/3292500.3330787}}


\bibitem[Zhao et~al\mbox{.}(2017)]%
        {zhao2017uplift}
\bibfield{author}{\bibinfo{person}{Yan Zhao}, \bibinfo{person}{Xiao Fang},
  {and} \bibinfo{person}{David Simchi-Levi}.} \bibinfo{year}{2017}\natexlab{}.
\newblock \showarticletitle{Uplift modeling with multiple treatments and
  general response types}. In \bibinfo{booktitle}{\emph{Proceedings of the 2017
  SIAM International Conference on Data Mining (SDM)}}.
  \bibinfo{pages}{588--596}.
\newblock
\href{https://doi.org/10.1137/1.9781611974973.66}{doi:\nolinkurl{10.1137/1.9781611974973.66}}


\bibitem[Zhong et~al\mbox{.}(2022)]%
        {zhong2022descn}
\bibfield{author}{\bibinfo{person}{Kailiang Zhong}, \bibinfo{person}{Fengtong
  Xiao}, \bibinfo{person}{Yan Ren}, \bibinfo{person}{Yaorong Liang},
  \bibinfo{person}{Wenqing Yao}, \bibinfo{person}{Xiaofeng Yang}, {and}
  \bibinfo{person}{Ling Cen}.} \bibinfo{year}{2022}\natexlab{}.
\newblock \showarticletitle{{DESCN}: Deep Entire Space Cross Networks for
  Individual Treatment Effect Estimation}. In
  \bibinfo{booktitle}{\emph{Proceedings of the 28th ACM SIGKDD Conference on
  Knowledge Discovery and Data Mining}}. \bibinfo{pages}{4612--4620}.
\newblock
\href{https://doi.org/10.1145/3534678.3539198}{doi:\nolinkurl{10.1145/3534678.3539198}}


\end{thebibliography}

\end{document}